\documentclass[11pt]{article}

\usepackage[final]{acl}
\usepackage{xurl}
\usepackage{times}
\usepackage{latexsym}
\usepackage{algpseudocode}
\usepackage[T1]{fontenc}

\usepackage[utf8]{inputenc}

\usepackage{microtype}

\usepackage{inconsolata}
\usepackage{amsmath}
\usepackage{graphicx}
\usepackage[table,xcdraw]{xcolor}
\usepackage{xcolor}
\usepackage{algorithm}

\title{CAAFC: \underline{C}hronological \underline{A}ctionable \underline{A}utomated \underline{F}act-\underline{C}hecker for misinformation / non-factual hallucination detection and correction}

 \author{Islam Eldifrawi \and Shengrui Wang \and Amine Trabelsi \\
         @usherbrooke.ca}

\begin{document}
\maketitle
\begin{abstract}
With the vast amount of content uploaded every hour, along with the AI generated content that can include hallucinations, Automated Fact-Checking (AFC) has become increasingly vital, as it is infeasible for human fact-checkers to manually verify the sheer volume of information generated online. Professional fact-checkers have identified several gaps in existing AFC systems, noting a misalignment between how these systems operate and how fact-checking is performed in practice. In this paper, we introduce CAAFC (Chronological Actionable Automated Fact-Checker), a framework designed to bridge these gaps. It surpasses SOTA AFC and hallucination detection systems across multiple benchmark datasets. CAAFC operates on claims, conversations, and dialogues, enabling it not only to detect factual errors and hallucinations, but also to correct them by providing actionable justifications supported by primary information sources. Furthermore, CAAFC can update evidence and knowledge bases by incorporating recent and contextual information when necessary, thereby enhancing the reliability of fact verification.
\end{abstract}

\section{Introduction}
Researchers have been investigating methods to produce AFC and hallucination detection systems, leveraging approaches from NLP, machine learning, knowledge representation, and database systems to predict the veracity of claims \cite{guo2022survey}.
\begin{figure*}
    \centering
    \includegraphics[width=10cm]{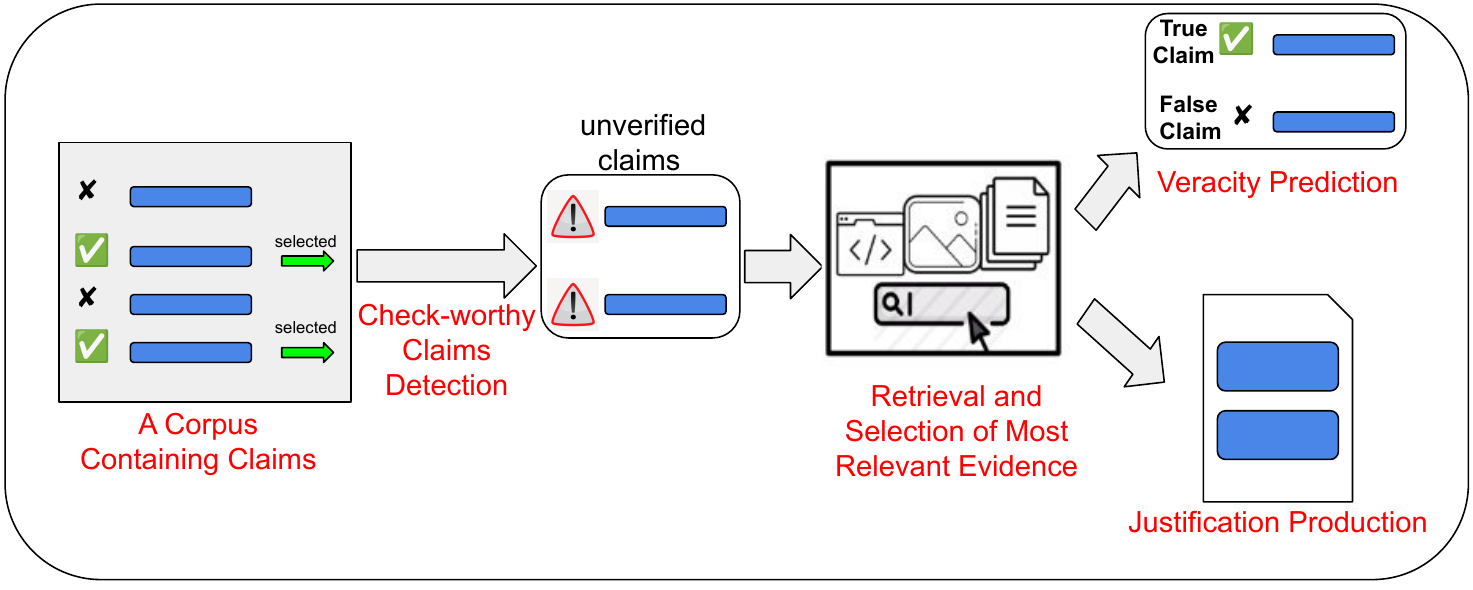}
    \caption{The classical AFC pipeline courtesy of \cite{guo2022survey}}
    \label{AFC_PI}
\end{figure*}

AFC pipelines use LLMs that start by detecting “check-worthy” claims, then they retrieve relevant evidence (information used as reference to judge the claim veracity) from trusted sources for each claim, then each claim is deemed true, false, or unverifiable, generating a human-readable explanation as Shown in Figure \ref{AFC_PI}. Hallucination detection systems aim to identify hallucinations in the outputs of LLMs, and mostly they rely on prompting LLMs to evaluate the generated content. However, many challenges are still present in both fields. We summarize the most pressing of these challenges in the form of the following four research questions:

\textbf{RQ1:} How can we obtain up-to-date evidence to judge claim veracity? Can we use this new evidence and update the knowledge of our databases?

\textbf{RQ2:} Professional fact-checkers follow different processes from most AFC systems. Does explicitly following fact-checkers' methodologies lead to improved fact-checking performance?

\textbf{RQ3:} Can the tasks of non-factual hallucination detection and automated fact-checking be unified, enabling the use of one detection and correction system for both tasks?

\textbf{ RQ4:} Given access to up-to-date evidence, is the use of large-scale frontier LLMs as fact-checkers still necessary? Alternatively, can AFC be decomposed into smaller sub-tasks that can be addressed by smaller quantized LLMs while preserving the performance?

\textbf{RQ5:} While most existing fact-checking systems focus primarily on claim verification, their generated justifications often lack explicit generation structure or rules. How can we generate high-quality, structured justifications that enhance transparency and foster user trust? 

\citet{10.1145/3706598.3713277} summarize the key differences between how professional fact-checkers conduct their work and how most AFC systems operate. One of the most significant distinctions lies in evidence retrieval, as professional fact-checkers prioritize primary / original sources such as official statements, and photos or videos documenting an alleged event. In contrast, most AFC systems rely heavily on secondary sources, such as media outlets reporting on an event, introducing interpretative bias rather than presenting raw evidence. In addition, professional fact-checkers provide explicit justifications for their choice of information sources as part of a robust editorial process. They also emphasize the importance of temporal grounding, ensuring that all retrieved evidence is accompanied by a timestamp, as factual correctness is inherently chronological. 

Another key distinction lies in how the veracity of claims is justified. For professional fact-checkers, the most critical property of an explanation is traceability \cite{10.1145/3706598.3713277}. This requires the inclusion of direct links to the evidence sources. The importance of providing supporting links for actionable explanations has also been emphasized by \citet{kotonya2020explainable}

Actionability was evaluated by \citet{eldifrawi-etal-2025-fingract} in their FinGrAct framework; however, their system itself does not generate actionable explanations. Specifically, actionable explanations are those that identify factual errors in false claims, provide explicit corrections, and cite the evidence sources that support these corrections. In this work, we generate actionable explanations using feedback from FinGrAct as an evaluation framework to assess the explanations quality. \textbf{Our contributions 
can be summarized as follows:}

\textbf{C1:} \textbf{To address RQ1}, CAAFC extract timestamped, ordered and updated evidence from the web. Using CAAFC, we introduce a novel method for detecting outdated or missing knowledge in databases, and for correcting this knowledge based on newly retrieved, chronological evidence. Furthermore, by incorporating chronology into the evidence, this ensures that the explanations reflect the \textbf{correct sequence of events}.

\textbf{C2:} \textbf{To address RQ2}, CAAFC follows the professional fact-checkers workflow in obtaining the evidence from its original primary source. For instance, consider a claim in which Trump tweeted that the the American economy state is great. Trump's Twitter account is preferred over news outlets reporting this event, as it serves as the primary source. By prioritizing the \textbf{original} sources of information, CAAFC avoids having biased evidence from secondary sources. 

\textbf{C3:} \textbf{To address RQ3}, CAAFC reformulates fact-checking and non-factual hallucination detection into a single task for misinformation detection and correction. 

\textbf{C4:} \textbf{To address RQ4}, CAAFC outperforms SOTA models (like GPT-4, GPT-5.2, Claude 3.5 Sonnet, and LLAMA3.1-70B) across multiple benchmark datasets in both tasks, with quantized Gemma3-27B demonstrating that dividing the complex task of misinformation detection and correction into simpler sub-tasks leads to decent fact verification.

Furthermore, We evaluate three quantized LLMs of varying sizes (Gemma3-27B, LLAMA3.3-70B, and GPT-OSS-120B) as CAAFC underlying LLMs. Remarkably, when provided with detailed, chronological evidence, the smaller Gemma3-27B model matches or surpasses the performance of much larger models, including GPT-OSS-120B and LLAMA3.3-70B, across multiple datasets in terms of accuracy and macro F1-score. This superior efficiency comes with a significant reduction in resource requirements: Gemma3-27B has over 4× fewer parameters than GPT-OSS-120B and 2.6× fewer than LLAMA3.3-70B, while requiring only 17 GB of GPU memory. These results highlight that, with chronologically ordered evidence, smaller LLMs can deliver high-performance fact-checking at a fraction of the computational cost.

\textbf{C5:} \textbf{To address RQ5}, CAAFC introduces many architectural changes to AFC and non-factual hallucination detection tasks. It adds new modules, such as the actionable justifier, actionability evaluator, and justification revisory, while replacing classical components like the claims check-worthiness detector with a claims extractor–segmentor that tries to extract as many claims as possible as shown in Figures \ref{AFC_PI} and \ref{pipe}. 

\section{Related Work}
In this section, we highlight other existing fact-checking systems and pipelines.
\citet{rothermel-etal-2024-infact} pipeline (INFACT) starts by enriching the claim with metadata (author and source URL) and prompting an LLM to reformulate it accordingly. INFACT then generates ten questions using manually curated in-context examples. For each question, the system retrieves evidence. Each retrieved document is embedded using the gte-base-en-v1.5 model and ranked by semantic similarity. INFACT iteratively evaluates the results, prompting the LLM to answer the question. After processing all questions, the LLM determines the claim's veracity. They achieved 0.72 accuracy on AVeriTeC. 


The Loki \cite{li-etal-2025-loki} pipeline comprises five components: a Decomposer that extracts individual claims, a Checkworthiness Identifier that filters non-worthy content, a Query Generator that formulates retrieval queries, an Evidence Retriever that collects supporting information, and a Claim Verifier that assesses the evidence and presents relevant snippets. This pipeline is optimized to minimize the number of steps and model calls, and is released as an open-source system that can operate with any LLM as its underlying model.

Most of the previous work primarily focuses on analyzing the strengths and weaknesses of LLMs across a broad range of tasks, without addressing the lack of chronological reasoning in the retrieved evidence. In particular, their reliance on SERP- or Serper-based is susceptible to retrieving evidence from secondary sources that add their own interpretation, their retrieval does not account for temporal consistency, nor do they tackle the challenge of generating high-quality justifications.
\section{Methodology}
\begin{figure*}
    \centering
    \includegraphics[width=11.7cm]{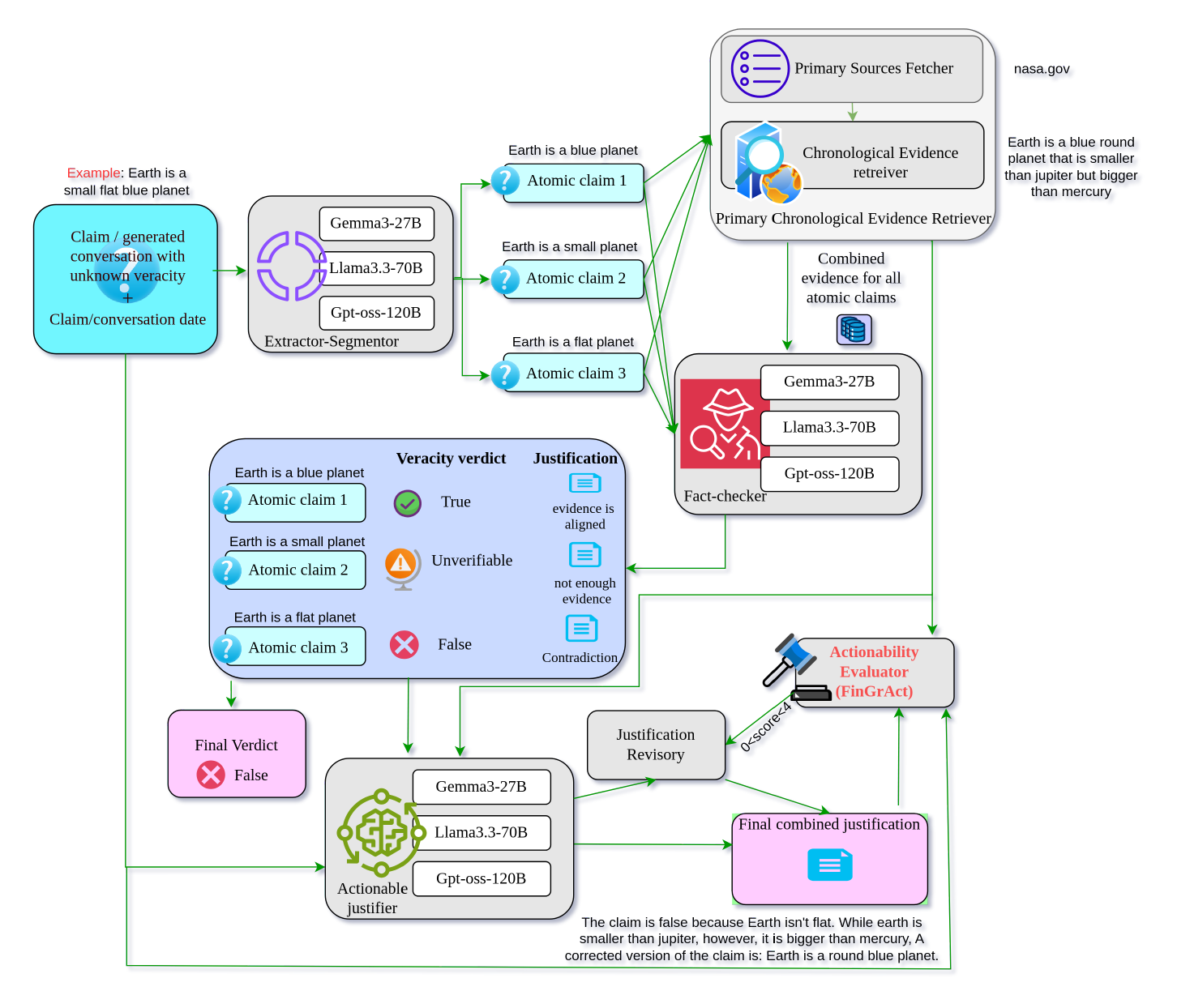}
    \caption{\textbf{CAAFC pipeline}. Mainly, CAAFC uses the quantized Gemma3-27B as its backbone, however, it can also use any LLM like llama3.3-70B or GPT-OSS120B. The example here is simplified and for illustration purposes. More elaborated examples are in the Appendix}
    \label{pipe}
\end{figure*}

CAAFC is a module based frame-work with six modules using quantized LLMs as shown in Figure \ref{pipe}. The modules are colored in gray. The input to CAAFC van be a claim that needs its veracity to be verified or a conversation that needs to be verified. The following paragraphs will discuss the six modules:

\textbf{The Extractor Segmentor:} The Extractor Segmentor is responsible for claim extraction and decomposing long and complex claims or conversations into simpler atomic claims that are easier to process and verify. This step is particularly important for claims containing multiple statements. For example, the claim “Earth is a blue, flat star” can be decomposed into three atomic claims;Earth is blue, Earth is a star, and Earth is round. \citet{mitra2025factlens} demonstrated that using atomic claims leads to more precise identification of inaccuracies, and greater transparency during evidence retrieval. Therefore, we base our prompt adopting their approach to segment claims into atomic units. \textcolor{black}{\textbf{We extract all possible claims, unlike what happens in the classical AFC pipelines, which only focus on check-worthy claims.}} The extractor segmentor prompt is in Figure \ref{seg_prmpt} in Appendix \ref{prmpts}.

\textbf{The Primary Chronological Evidence Retriever:} This module processes the list of atomic claims, together with their corresponding date, to retrieve reliable and time-relevant information. \textcolor{black}{\textbf{Unlike classical AFC, we direct our evidence retrieval towards the primary sources that should be checked to verify the list of the atomic claims.}} Fact-checkers  don't cite, for example, media citing source, they try to find the most primary, most original source of data \cite{10.1145/3706598.3713277}. For example, consider the claim made by Donald Trump in 2025 that the Biden administration spent millions on making mice transgender. Secondary sources such as CNN (often characterized as left-leaning), IMDB, and other outlets reported that the accusation was false. In contrast, primary sources, including official expenditure records published on WhiteHouse.gov, were interpreted as supporting the claim. To mitigate the risk of retrieving evidence from potentially biased sources, we introduce directive search, which guides retrieval toward primary-source evidence. \textcolor{black}{\textbf{The module searches for chronological knowledge encompassing both the time when the claim was made and the present using the fetched primary sources, thereby providing as much contextual evidence as possible which is different from classical AFC.}} We use the publicly available Google AI Mode \footnote{\url{https://Google.com/ai}}, which summarizes search results from Google using Gemini-Pro-3. The retrieved information is ranked according to the criteria described in Appendix \ref{g_rank}, ensuring that the selected sources meet high standards of relevance, quality, and authority. The query used is\textit{ "I need timestamped information about the following list of claims \{atomic claims list\} on the following date \{date\}. Start by checking the following sources: \{primary\_sources\_list\} ".} The output consists of chronologically ordered evidence, with information relevant to the claim's date as well as updated knowledge reflecting the current date supported with the sources links. Google validates its AI mode summarization as shown in Appendix \ref{Google_val}.

\textbf{The Fact-Checker:} The fact-checker determines whether each atomic claim is supported, contradicted, or unverifiable based on the retrieved evidence, producing a veracity verdict along with a justification for each prediction using the prompt shown in Figure \ref{fact_check_prmpt} in Appendix \ref{prmpts}. After evaluating all atomic claims, the final verdict for the overall claim is derived using Algorithm \ref{alg:final_verdict}.

\begin{algorithm}[t]
\caption{Final Veracity Verdict Aggregation}
\label{alg:final_verdict}
\begin{algorithmic}[1]
\Require Set of atomic claims $\mathcal{A} = \{a_1, a_2, \dots, a_n\}$ with veracity labels
\Ensure Final veracity verdict $V$

\State $V \gets \text{true}$

\For{each atomic claim $a_i \in \mathcal{A}$}
    \If{$a_i$ is \textbf{false}}
        \State \Return $\text{false}$
    \ElsIf{$a_i$ is \textbf{unverifiable}}
        \State $V \gets \text{unverifiable}$
    \EndIf
\EndFor

\State \Return $V$
\end{algorithmic}
\end{algorithm}

\textbf{The Actionable Justifier:} This module is new to the AFC pipeline and integrates the claim, the retrieved evidence, and the atomic claims with their respective justifications to generate a combined actionable justification. In the context of AFC, actionability refers to producing an explanation that not only highlights the specific errors present in the claim but also provides corrections for those errors and its sources. The prompt used in this module is shown in Figure \ref{act_just_prmpt} in Appendix \ref{prmpts}.

\textbf{The Actionability Evaluator:} To ensure that the generated justification is actionable, this module evaluates its quality using FinGrAct \cite{eldifrawi-etal-2025-fingract}. The interaction between this module, the justification revisory module, and the actionable justifier module ensures actionability is produced within the justifications. In FinGrAct, a score of zero mostly indicates that the claim is true and does not require an actionable explanation. FinGrAct evaluates the actionability of a justification based on three criteria. The first criterion assesses whether the justification highlights all factual errors in the claim: a score of 0 indicates that it does not detect any errors, 1 indicates partial detection, and 2 indicates that all errors in the claim are correctly identified. The second criterion evaluates whether the justification corrects the identified errors. A score of 0 means no corrections are provided, 1 means partial correction, and 2 means that all errors are fully corrected. The third criterion measures whether the information sources cited in the evidence are functional, relevant, and supportive of the justification. A score of 0 indicates that the links are non-functional, 1 that they are functional, 2 that they are relevant, and 3 that they fully support the justification. The total FinGrAct score is obtained by summing the scores from these criteria, resulting in a maximum score of 7 points. 

Therefore, if the justification's actionability score is at least 4 out of 7 points, the justification is considered to have an acceptable degree of actionability. If the justification fails to meet this threshold, the module identifies its weaknesses and provides feedback to the justification revisory, prompting an enhanced and more actionable revision. This refinement process aims to achieve a score of 4 points or more on the actionability scale. 

\textbf{Justification Revisory:} This new module refines the generated justification by incorporating feedback from FinGrAct, with the goal of improving its actionability. Specifically, it enhances the justification by adding missing corrections and highlighting previously unaddressed factual errors. We mainly employ Gemma 3–27B in this module to reduce computational overhead while maintaining performance. The prompt for the justification revisory module is in Figure \ref{just_rev} in Appendix \ref{prmpts}.

The final outputs of the CAAFC framework are highlighted in pink in Figure \ref{pipe}. To compare the performance across models of different scales, we experimented with the three open-source LLMs shown in Figure \ref{pipe}, however, we prefer the quantized Gemma3-27B as the backbone of CAAFC as it is smaller and efficient. 

\section{Datasets}
To evaluate the performance of CAAFC, we used four benchmark datasets. For each dataset, we relied solely on the test set, since our approach is based on zero-shot CoT prompting, therefore, does not require access to the training data. 

The first dataset, AVeriTeC  \cite{schlichtkrull-etal-2024-automated}, contains 500 real-world claims in its test set. Each claim is accompanied a fine-grained veracity label (122 supported claims, 305 refuted, 35 not enough evidence, and 38 conflicting evidence).
We merged the two classes "not enough evidence" and "conflicting evidence" into a single class labeled "unverifiable". Dataset labels aggregation is to unify the output of CAAFC, however, CAAFC justifications provide fine-grained details on its classification. More details on aggregation and its effect are in Appendix \ref{lab_agg}. 

The second dataset, CoverBench \cite{Jacovi2024CoverBenchAC}, 
contains 733 claims in the testset (330 labeled as false and 403 as true). Each entry includes the claim, its unprocessed evidence, the label, the claim's source, and its domain. The primary purpose of this dataset in our evaluation is to assess the robustness of our framework in handling complex, and domain-diverse claim verification.

The third dataset, FACTors \cite{10.1145/3726302.3730339}, contains a collection of recent claims spanning a broad chronological range. Notably, 196 of these claims emerged after the release dates of models such as Llama3.3-70B and Gemma3-27B, making this dataset particularly suitable for evaluating our framework's ability to verify claims that the underlying LLMs have not been previously exposed to as shown in Appendix \ref{af_ll}. 
The FACTors test set comprises 5000 claims, categorized as follows: 221 true, 3735 false, 632 misleading, 264 partially true, and 148 unverifiable claims. Following our classification strategy, we merged the misleading, and partially true categories into the false class.


The final dataset is DiaHalu \cite{chen-etal-2024-diahalu}, a challenging dialogue-based benchmark featuring longer conversations and multiple categories of hallucinations. We focus exclusively on the factual and non-factual dialogues; we have 122 and 123 conversations respectively. We also experimented with HaluEval dataset that focuses on non-factual hallucination detection in dialogues without any sampling  in Appendix \ref{H_EV}.
\section{Experiments and Results} 
\subsection{Evidence Evaluation}
In this experiment, \textbf{we aim to evaluate the quality of the evidence retrieved from Google.} In the literature, the proxy-reference evaluation method \cite{huang-zhang-2021-evaluation,akhtar2024ev2r} is commonly used for this purpose. Their approach assesses evidence based on two criteria: (1) the veracity verdict, where a correct verdict implies that the retrieved evidence is valid, and (2) the similarity between the retrieved evidence and the reference evidence. However, this method implicitly assumes that the reference evidence is always optimal, overlooking the possibility that it may be outdated or lacking contextual information that was unavailable at the time it was collected.

To identify instances where the retrieved evidence may surpass the dataset evidence-for example, when the latter is outdated or lacks sufficient context-we prompted three LLMs to generate independent veracity verdicts. Specifically, we used Llama 3.3-70B, Gemma 3-27B, and GPT-OSS-120B. When all LLMs produced the same veracity verdict that contradicted the dataset's label given the retrieved evidence (e.g., a claim labeled as true in the dataset but predicted as false by all LLMs), we conduct a comparative analysis between our retrieved evidence and the evidence provided in the dataset. The three LLMs are just used as a candidate indicator of problematic instances that are thoroughly analyzed in Appendix \ref{EA} .

We analyzed all the instances where LLMs agree among themselves on a certain veracity label but they contradict with the dataset label. We found that this contradiction is either because of outdated information, or missing context in the dataset evidence information. \textbf{For instance, consider the claim:} \textcolor{black}{“\textit{The U.S. Surgeon General previously said that masks are not effective in preventing the general public from catching coronavirus.}”} \textbf{In the AVeriTeC  dataset, the associated evidence is:} \textcolor{black}{\textit{“There is a 79\% reduction made in certain situations, but not effective in certain other situations." \textbf{SOURCE:} https://thefederalist.com/2020/10/29/these-12-graphs-show-mask-mandates-do-nothing-to-stop-covid/}}. Based on the dataset evidence collected in 2020, the dataset veracity label is true. \textbf{While the Google evidence -with information source in 2021- is :} \textcolor{black}{"\textit{During the early stages of the COVID-19 pandemic in February and March 2020, the U.S. Surgeon General issued statements discouraging the general public from buying and wearing masks. This guidance was later reversed, and officials explained that the initial recommendation was based on the need to preserve limited supplies of medical masks for healthcare workers." \textbf{SOURCE}: https://www.latimes.com/science/story/2021-07-27/timeline-cdc-mask-guidance-during-covid-19-pandemic}}. Based on the updated Google evidence, all three LLMs deemed the claim as false. More detailed examples on this pattern are found in Appendix \ref{EA}. We refer to this phenomenon as an “evidence chronological mismatch.”

Another example from DiaHalu showing that the dataset evidence was missing the context needed to verify the conversation: \textcolor{black}{\textit{[A1]: Do you like Michael Jackson? [B1]: I do. He wrote Dancing on the Dream which is one of my favorite songs.  [A2]: Yes, he was very talented. He also won Grammy of the Year. [B2]:"Yes, Healing the World is one of my favorite songs."}} \textbf{The dataset evidence:} \textcolor{black}{\textit{Michael Jackson wrote Dancing the Dream. }} Based on this evidence, the conversation was labeled factual. \textbf{Google evidence: } \textcolor{black}{\textit{Michael Jackson wrote a book of poetry titled Dancing the Dream, published in 1992. It is not a song."}} All LLMs marked this conversation as a hallucination. To further validate our analysis, three human annotators were provided with 100 these instances. There was substantial agreement among them that the Google evidence was better as shown in Appendix \ref{hum_ann}.

In AVeriTeC , 57 instances exhibited the evidence chronological mismatch pattern. Similarly, 73 instances were identified in CoverBench, and 245 instances in FACTors. 
\textbf{All such instances will be removed during the data cleaning process, and experiments will be conducted both with and without data cleaning.} To automate the detection of such instances, further details are in Appendix \ref{evd_update}. In addition, we used CAAFC on the datasets evidence in Experiment \ref{no_ev_ret} to test its performance without the evidence retriever and without any data cleaning. 
\subsection{Evaluating CAAFC based on the datasets evidence and prior to data cleaning then after data cleaning}
\label{no_ev_ret}
In this experiment, there are two stages. \textbf{First, we remove the evidence retrieval module and we use the evidence in the datasets (No data cleaning is performed). We compare our approach}, which divides AFC into smaller subtasks handled by a quantized, smaller Gemma3-27B, \textbf{against larger, SOTA models that process the entire task end-to-end}. This experiment \textbf{shows how much improvement comes from better data after data cleaning versus the framework itself without data cleaning}. 

On AVeriTeC , CAAFC achieved an accuracy of 0.780 using zero-shot prompting with Gemma 3–27B, outperforming InFact \cite{rothermel-etal-2024-infact}  that reachd 0.720 accuracy and the fine-tuned LLaMA 3.1-70B model reported by \citet{yoon2024hero}, which achieved 0.752 accuracy. For CoverBench, CAAFC obtained an accuracy of 0.729 and a macro F1-score of 0.712 using the smaller quantized Gemma 3-27B, compared to 0.689 achieved by the SOTA Claude 3.5 Sonnet model. On FACTors, CAAFC recorded an accuracy of 0.870 using Gemma 3-27B, while their fine-tuned RoBERTa model reached 0.849 accuracy. 

Overall, these results indicate that decomposing the complex task of fact-checking into simpler subtasks within a structured pipeline, and addressing them with smaller, efficient LLMs can yield comparable or sometimes better performance to that of larger models applied to the task end-to-end. This is consistent with the findings of \citet{xing-etal-2025-llmsr, fysikoudi2025active, xu-etal-2025-team}

In the second stage of this experiment, we removed all instances exhibiting the first error pattern, in which the three LLMs contradicted the dataset label due to a chronological mismatch between the Google AI evidence and the dataset evidence, or missing content in the dataset evidence. the aim of this stage is to study the effect of data cleaning.

We relied on the dataset evidence and labels after the data cleaning and we compare the performance before and after data cleaning. On AVeriTeC, the accuracy increased to 0.88, representing an improvement of +0.10. On CoverBench, the accuracy rose to 0.81, corresponding to a gain of +0.081. On FACTors, the accuracy improved to 0.914, reflecting an increase of +0.044. It is worth noting that in this experiment, we adhered to the performance metrics used by the SOTA baselines of each dataset to ensure a fair comparison with CAAFC.

\subsection{Evaluating CAAFC on different datasets with different underlying LLMs relying on Google evidence}
\label{gogogo}
\begin{table*}
\centering
\resizebox{12cm}{!}{%
\begin{tabular}{c|cc|cc|cc|}
\cline{2-7}
 & \multicolumn{2}{c|}{GEMMA3-27B} & \multicolumn{2}{c|}{LLAMA3.3-70B} & \multicolumn{2}{c|}{GPT-OSS-120B} \\ \cline{2-7} 
 & \multicolumn{1}{c|}{\begin{tabular}[c]{@{}c@{}}Macro\\ F1-score\end{tabular}} & Accuracy & \multicolumn{1}{c|}{\begin{tabular}[c]{@{}c@{}}Macro\\ F1-score\end{tabular}} & Accuracy & \multicolumn{1}{c|}{\begin{tabular}[c]{@{}c@{}}Macro\\ F1-score\end{tabular}} & Accuracy \\ \hline
\multicolumn{1}{|c|}{AVeriTeC } & \multicolumn{1}{c|}{\textbf{0.776}} & \textbf{0.894} & \multicolumn{1}{c|}{0.764} & 0.892 & \multicolumn{1}{c|}{0.741} & 0.893 \\ \hline
\multicolumn{1}{|c|}{Coverbench} & \multicolumn{1}{c|}{0.825} & 0.829 & \multicolumn{1}{c|}{0.854} & 0.855 & \multicolumn{1}{c|}{\textbf{0.858}} & \textbf{0.859} \\ \hline
\multicolumn{1}{|c|}{Factors} & \multicolumn{1}{c|}{0.693} & 0.919 & \multicolumn{1}{c|}{\textbf{0.708}} & \textbf{0.931} & \multicolumn{1}{c|}{0.540} & 0.774 \\ \hline
\end{tabular}%
}
\caption{Accuracies and Macro-F1 scores showing the performance of our framework \textbf{relying on Google chronological retrieved evidence} in multiple \textbf{cleaned} datasets and benchmarks.}
\label{framework_perf}
\end{table*}
In this experiment, the veracity prediction performance of CAAFC was evaluated using macro F1-score and accuracy \textbf{relying on Google evidence} on AVeriTeC , CoverBench, and FACTors datasets. We relied on the dataset labels as the ground truth, except in cases where an evidence chronology mismatch pattern was observed. All these instances were further analyzed and the LLMs consensus labels are found to be  more consistent. Furthermore, human evaluation was conducted on a subset of them. The results showed strong agreement among the annotators—across 100 random instances out of the total 375—that the Google-retrieved evidence was better, and that the label predicted by the LLM was more consistent with the majority vote of the human annotators in Appendix \ref{hum_ann}. We removed such instances and we experimented on the data after cleaning. \textbf{This experiment aims to measure the effect of having chronological evidence on classification performance, unlike previous experiments where it wasn't available.} 
The results in Table \ref{framework_perf} show a slight improvement over the previous experiment. 

Interestingly, Gemma 3-27B achieved performance comparable to much larger models, such as Llama 3.3-70B and GPT-OSS-120B, in claim veracity prediction. This observation suggests that model size alone is not the determining factor in fact verification; rather, the quality and chronology of the retrieved evidence play a more critical role.


When analyzing the errors and misclassifications, the most dominant pattern was when the LLMs mis-interprets the input.  We name this error pattern the 'LLM mis-interpretation / hallucination error'. As an example, here is this true-labeled claim: \textit{\textcolor{black}{"Canada Living standards in 2023 went down and people were worse off at the end of the last government than they were at the beginning of the government."}}. with this Google evidence: \textit{\textcolor{black}{" 
It is not clear which specific time period or measure of living standards was referred to. However,  the Bank of Canada reported that GDP per person fell for six consecutive quarters from mid-2022 until the end of 2023. "}} GPT labeled it as unverifiable and here is its justification: \textit{\textcolor{black}{\{'justification': "
The original statement lacks specificity regarding the time period and measure of living standards. 
"\}}} While Gemma labeled the claim as true because: \textit{\textcolor{black}{\{'justification':Both subclaims are supported by the evidence, showing a consistent decline in GDP per person. The Bank of Canada reports indicate a decrease in living standards, confirming that people were, on average, worse off at the end of the government's term compared to the beginning". 
\}}}
\begin{figure*}
    \centering
    \includegraphics[width=11cm]{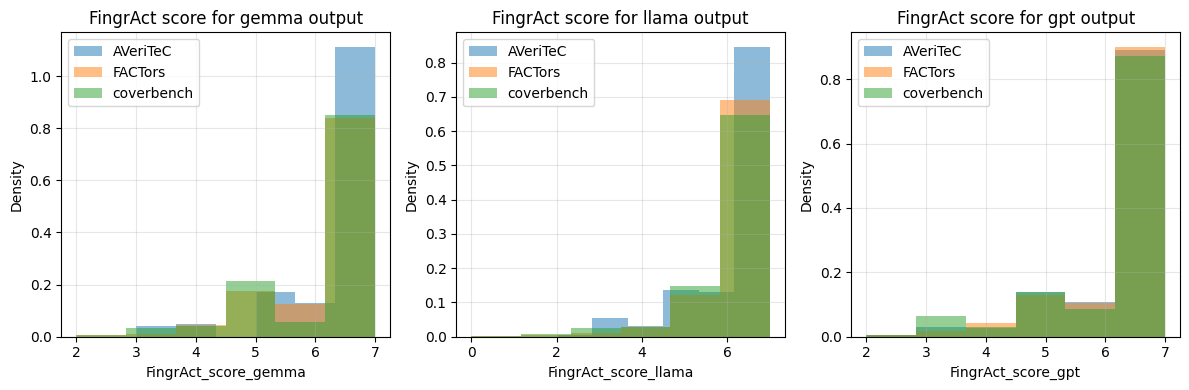}
    \caption{Density for actionability scores by FinGrAct}
    \label{hist}
\end{figure*}

\subsection{Generating high quality explanations with CAAFC}

In this experiment, we evaluated the degree of actionability of the actionable justifier's responses using FinGrAct. \citet{eldifrawi-etal-2025-fingract} reported a strong correlation between faithfulness, coherence, and actionability. This suggests that generating actionable justifications is likely to enhance their coherence and faithfulness, leading to reliable and comprehensible explanations. Upon examining the link score across the AVeriTeC , CoverBench, and FACTors datasets, we observed that all links retrieved by the Google were functional, relevant, and supportive of the corresponding evidence. Details of the FinGrAct scoring system and our chosen actionability threshold are in Appendix \ref{FinGrAct_scr}.

After applying FinGrAct to evaluate the actionability of the justifications produced by CAAFC, we observed that most responses were highly actionable, with the majority of FinGrAct scores ranging between 6 and 7 points out of 7. Figure \ref{hist} presents the histogram distributions for the three datasets
. To ensure comparability across datasets of varying sizes, we report density plots rather than raw frequencies. The majority of justifications that had actionability scores below 4 were for true claims as shown in Table \ref{perc_scr} in Appendix \ref{FinGrAct_scr} since true claims aren't actionable.


\subsection{Evaluating CAAFC against hallucination detection in conversations}
We evaluate CAAFC on dialogues of the  sampled DiaHalu dataset without any data cleaning, to assess the framework's ability to detect and correct hallucinations. 
For DiaHalu, after implementing system of \citet{chen-etal-2024-diahalu} using GPT-4 and their reported parameters, it achieved 0.473 macro F1-score in factual and non-factual dialogues detection, , while \citet{rathore2026temporal} reported 0.495 using Temporal Graph Networks. We used \citet{chen-etal-2024-diahalu} hallucination detection prompt and parameters on GPT-5.2 (more details in Appendix \ref{diaHalu_exp}) and it reported 0.682  macro F1-score. CAAFC using Gemma3-27B achieved 0.684 macro F1-score. This improvement over GPT-5.2 demonstrates the capability of CAAFC to both detect and correct non-factual hallucinations despite its smaller size. Detailed metrics and details are in Appendix \ref{diaHalu_exp}.

In this experiment, the Actionable Justifier, Justification Revisory, and Actionability Evaluator modules were disabled. The pipeline therefore consisted of only three components: the extractor-segmentor, the chronological evidence retriever, and the fact-checker. Each claim required three calls—one per active module. On average, the inference time for the extractor-segmentor using Gemma 3–27B was 1.5 seconds. The chronological evidence retrieval, implemented via Selenium on Google AI Mode, required approximately 3 seconds. The fact-checker module required an additional 1.5 seconds, resulting in a total average processing time of 6 seconds per claim. By comparison, GPT-5.2, implemented following the framework of \citet{chen-etal-2024-diahalu}, required an average of 3.5 seconds per claim. 

\section{Conclusion}
We introduce CAAFC, a framework designed to breakdown AFC into simpler subtasks. CAAFC consistently outperforms SOTA models across multiple datasets, and can be applied to detect and correct hallucinations in LLMs generations. 
In addition, CAAFC produces actionable, and chronological justifications, supported by reliable sources, thereby enhancing transparency, and explainability.

\section*{Limitations}
\begin{enumerate}
    \item \textbf{Language limitation:} CAAFC is currently designed and evaluated exclusively on English-language content. Its performance on claims, conversations, or evidence in other languages remains unexplored and may require multilingual reasoning capabilities. In future work, we will explore the usage of multi-linguage LLMs and perform evaluations on multi-lingual datasets.
    \item \textbf{Risks of automatically updating knowledge bases from web evidence: }A further concern arises from the political and ideological skew that may be implicitly encoded in search engines and web-scale language models, which CAAFC leverages as its primary evidence source. Because ranking algorithms and training data can over-represent state or corporate media, particular geopolitical blocs, or high-SEO partisan outlets, the retrieved evidence may systematically privilege certain narratives over others, especially on contentious topics such as elections, pandemics, or international conflicts.
    \item \textbf{Computational overhead:} The framework may require multiple LLM calls per instance, which can increase inference latency. However, it offers a substantial reduction in GPU memory usage. CAAFC requires only 17 GB of GPU RAM using the quantized Gemma3-27B LLM, whereas GPT-5.2 would require significantly more. In future work, we plan to explore fast APIs for the evidence retrieval component and upgrade our hardware for higher compute power to further reduce overall response time. In addition, our experiments were conducted using an NVIDIA Quadro RTX 8000 GPU. Since the hardware specifications used by OpenAI for GPT-5.2 is not publicly disclosed, a direct comparison of inference times may not be entirely equitable.

\end{enumerate}

\bibliography{custom}

\appendix

\section{Appendix}
All experiments were conducted on a Quadro RTX 8000 GPU and all models were quantized (Q4\_K\_M) to address GPU RAM limitations. Temperature is set to zero for all the involved LLMs.
\subsection{FinGrAct Scoring}
\label{FinGrAct_scr}
A score of 0 indicates that the claim is true, and no supporting links were provided by the Google. When the score equals 3, three main scenarios may occur: (1) supporting links for a true claim are provided, (2) the error detection score is 2 while the error correction score is 1 with no links provided, or (3) the error correction score is 2 while the error detection score is 1 with no links provided. For scores above 3, the link score remains 3, and the increase arises from improvements in error detection and/or error correction. This observation justifies setting our acceptance threshold at 4.
Table \ref{perc_scr} shows that the majority of justifications that took actionability scores less than 4 are for true claims.

\subsection{Ablation studies}
The comparisons between CAAFC and the aforementioned state-of-the-art models are based on accuracy and macro F1-score of the predicted verdicts or labels. Therefore, removing modules designed to enhance justification quality—such as the actionable justifier, justification revisory, and actionability evaluator—would not substantially affect overall classification performance. Nevertheless, these modules remain important for producing high-quality justifications.

The Fact-Checker module is central to the AFC process and therefore indispensable. Moreover, the extractor–segmentor produces outputs that feed into the Fact-Checker, the final verdict algorithm, and the actionable justifier; removing it would require substantial redesign of CAAFC’s architecture. Importantly, the extractor–segmentor gives CAAFC an advantage over some frontier LLMs by extracting claims that —including those that present subjective opinions as facts and those inferred from paragraphs—as demonstrated in Appendix \ref{diaHalu_exp}.

We evaluated the impact of the evidence retrieval component by removing it and measuring CAAFC’s performance in its absence in experiment \ref{no_ev_ret}. In addition, we evaluated the impact of removing the primary source fetcher by measuring the resulting accuracy and F1-score across the datasets. No significant changes were observed in either metric. However, we argue that this component remains essential, as Google search may rely on secondary sources such as blogs or platforms like Reddit, which can contain biased or unverified information. Therefore, we retain this module to ensure greater evidence reliability and robustness. Furthermore, this component enhances the reproducibility of the evidence when primary sources exist, since the retrieved evidence originates from the same source.

It is worth noting that incorporating chronological evidence retrieved from Google led to consistent accuracy improvements across all datasets between Experiment \ref{no_ev_ret} after data cleaning and Experiment \ref{gogogo} using Gemma3-27B in CAAFC. On AVeriTeC , accuracy increased from 0.880 to 0.8940; on CoverBench, from 0.810 to 0.829; and on FACTors, from 0.914 to 0.919. These results indicate that incorporating chronology into the evidence retrieval process positively impacts fact-verification accuracy. FACTors is the most recent dataset among the three, and therefore likely required fewer evidence updates compared to AVeriTeC  or CoverBench.
\begin{table}
\resizebox{\columnwidth}{!}{%
\begin{tabular}{c|cc|cc|cc|}
\cline{2-7}
 & \multicolumn{2}{c|}{AVeriTeC } & \multicolumn{2}{c|}{CoverBench} & \multicolumn{2}{c|}{FACTors} \\ \cline{2-7} 
 & \multicolumn{1}{c|}{true} & non-true & \multicolumn{1}{c|}{true} & non-true & \multicolumn{1}{c|}{true} & non-true \\ \hline
\multicolumn{1}{|c|}{\begin{tabular}[c]{@{}c@{}}Percentage of\\  claims\end{tabular}} & \multicolumn{1}{c|}{0.820} & 0.180 & \multicolumn{1}{c|}{0.778} & 0.222 & \multicolumn{1}{c|}{0.527} & 0.473 \\ \hline
\end{tabular}%
}
\caption{Percentage of true claims vs non-true claims whose justifications are evaluated by FinGrAct and has actionability scores less than 4.}
\label{perc_scr}
\end{table}

\subsection{Prompts}
\label{prmpts}

In this section, we present the prompts used by the individual modules that collectively compose the CAAFC framework.
\begin{figure}
\fcolorbox{blue}{gray!15}{%
        \begin{minipage}{\columnwidth}
\scriptsize
We have a transcript that has many claims. Extract all possible claims whether verifiable or not. To make the fact-checking task simpler, we break complex claims down into simpler, atomic sub-claims. \\Note that atomic sub-claims refer to unit claims within the original claim, that refer to a single concept that can be independently verified without having to refer to the original claim. Verification of the sub-claims should not require aggregation of facts or multi-hop reasoning over concepts. However, the sub-claim should have all the contextual information preserved from the original claim. Your task is to break down a claim into atomic sub-claims for fact checking only if needed. If the original claim itself is a unit claim, do not break it down.
\\Note how each sub claim contains atomic information to fact check and is brief, yet is contextualized with all the information needed from the original claim.
Now find the sub claims from the following text.\\

Text: \{text\}\\

Sub\_Claims: < your output in form of a list >
 \end{minipage}%
}
    \caption{prompt used for extracting claims and segmenting them into atomic claims}
    \label{seg_prmpt}

\end{figure}

\begin{figure}
\fcolorbox{blue}{gray!15}{%
        \begin{minipage}{\columnwidth}
\scriptsize
\textbf{Role:}
You are a chronology-aware automated fact-checking agent.

\textbf{Goal:}
Given a claim, identify the minimum set of primary authoritative sources that can directly verify or falsify it.

1. Prefer sources that:

\begin{itemize}
    \item Produce original data
    \item Hold legal, scientific, or operational authority
    \item Are required for real-world systems to function
\end{itemize}

2. Rank sources by authority strength, not popularity.

3. Distinguish between measurement authority, regulatory authority, and theoretical authority.

4. Avoid secondary explainers, media articles, and encyclopedias.

5. Your output should be a list of primary sources and the justification for their selection.
the claim:\\
\{claim\}
 \end{minipage}%
}
    \caption{Prompt used for extracting primary sources.}
    \label{prim_srcs_prmpt}

\end{figure}

\begin{figure}
\fcolorbox{blue}{gray!15}{%
        \begin{minipage}{\columnwidth}
\scriptsize
Role: You are a fact-checking assistant. Your task is to analyze a claim and determine the truthfulness of its sub-components based on the provided evidence.\\

You are given a list of sub-claims, and the evidence which is the reference information used to judge the claim or dialogue veracity.
Follow these steps carefully:\\
1. Compare each subclaim with the provided evidence.\\
2. Check whether the evidence supports, contradicts, or does not mention the subclaim. Base your judgment only on the given evidence. Do not rely on prior knowledge.\\
3. Assign one of three labels to each subclaim:

    - "true" if The evidence directly supports the subclaim.
    
    - "false" if The evidence directly contradicts the subclaim.
    
    - "unverifiable" if The evidence is insufficient, unclear, or unrelated to verify the subclaim.\\
4. Your output is a json object containing only three fields which are the subclaim, label, and the explanation. Here is an example of the output format:\\
\{"subclaims": [\{"text": "First subclaim here.","label": "true | false | unverifiable","justification": "Brief explanation of how the evidence supports, contradicts, or fails to verify this subclaim."\}, \{"text": "Second subclaim here.",label": "true | false | unverifiable","justification": "Brief explanation."\}]\}\\
5. Here is an example of the whole process:

   sub-claims: ['Paris is the capital of Germany', 'Paris has the Eiffel Tower.']
   evidence: Paris is the capital of France. The Eiffel Tower is located in Paris.
   
   output: \{"subclaims": [\{"text": "Paris is the capital of Germany.","label": "false","justification": "The evidence states Paris is the capital of France, not Germany."\}, \{"text": "Paris has the Eiffel Tower.","label": "true","justification": "The evidence confirms the Eiffel Tower is located in Paris."\}]\}\\
6. Do not add the instructions or anything from the prompt. Your output is restricted to the json format mentioned in step 4.\\

claim:\\
\{claim\}\\
\\
evidence:\\
\{evidence\}
 \end{minipage}%
}
    \caption{Prompt used by the fact-checker module}
    \label{fact_check_prmpt}

\end{figure}

\begin{figure}
\fcolorbox{blue}{gray!15}{%
        \begin{minipage}{\columnwidth}
\scriptsize
You are given a claim, evidence (reference information used to judge the claim), and a JSON object. The json object contains the claim divided into subclaims, a judgement for each subclaim: true, false, or unverifiable, and a justification for each subclaim based on the evidence.\\
Your task is to produce an actionable justification for the overall claim. Follow these instructions carefully:\\

1. Clearly reference which subclaims strengthen or weaken the claim\\
2. Explain how the evidence supports or contradicts the claim at a structural level\\
3. Highlight any gaps or uncertainties caused by unverifiable subclaims\\
4. Describe how false subclaims impact the credibility of the overall claim\\
5. Provides a corrected version of the whole claim.\\
6. Do not rewrite or re-evaluate the subclaims. Only analyze and integrate what is already in the JSON object.\\
7. Here is an example of the whole process:\\
   claim: Paris is the capital of Germany and it has the Eiffel Tower.\\
   evidence: Paris is the capital of France. The Eiffel Tower is located in Paris.\\
   json object: {"subclaims": [{"text": "Paris is the capital of Germany.","label": "false","justification": "The evidence states Paris is the capital of France, not Germany."}, {"text": "Paris has the Eiffel Tower.","label": "true","justification": "The evidence confirms the Eiffel Tower is located in Paris."}]}\\
   output: {"justification": "the claim has a factual error in the part where it says that Paris is the capital of Germany as Paris is in France. The corrected version of this claim is 'Paris is the capital of France and it has the Eiffel Tower.'"}\\
8. Do not add the instructions or anything from the prompt. Your output is restricted to the json format mentioned in step 7.\\

claim:
\{claim\}\\

evidence:
\{evidence\}\\

json object:
\{json\_object\}
 \end{minipage}%
}
    \caption{Prompt used by the actionable justifier module}
    \label{act_just_prmpt}

\end{figure}

\begin{figure}
\fcolorbox{blue}{gray!15}{%
        \begin{minipage}{\columnwidth}
\scriptsize
\textbf{Role:} You are an expert fact-checking editor. Your task is to improve the actionability of justifications by explicitly addressing missing errors and corrections identified in reviewer feedback.\\

\textbf{User prompt:}\\

You are given a justification explaining a verdict for a factual claim, and feedback that identifies unmentioned errors, missing corrections, or incomplete reasoning in the justification. Your task is to revise the justification so that it becomes more actionable and complete.\\

Specifically:\\

1.Explicitly add corrections that were missing but mentioned in the feedback.\\
2. Clearly highlight previously unmentioned errors in the original justification.\\
3.Preserve all correct reasoning already present.\\
4.Do not introduce new facts or evidence beyond what is implied by the feedback.\\
5. Use clear, concise, and structured language suitable for professional fact-checking reports.\\
6. Produce a single revised justification.\\
7. Ensure the reasoning clearly connects errors, corrections, final verdict.
8. Avoid meta-commentary (e.g., “the feedback says…”).

Original Justification:
\{justification\}

Feedback on Missing Errors and Corrections:
\{feedback\}
 \end{minipage}%
}
    \caption{Prompt for justification revisory}
    \label{just_rev}

\end{figure}

\subsection{Automation of evidence update}
\label{evd_update}

To fully automate the evidence inspection methodology, we zero-shot prompt LLAMA3.3-70B, Gemma3-27B, and GPT-OSS-120B using the prompt shown in Figure \ref{ev_compr}, in order to compare the evidence provided in the original datasets with the evidence retrieved via Google. We apply a majority-voting scheme over the predictions of the three LLMs to determine whether the original dataset evidence or the Google-retrieved evidence is superior, based on factors such as contextual relevance, temporal validity, and whether the evidence is up-to-date or outdated.

We focus on the examples where the chronological mismatch pattern (first pattern) was observed between the original evidence in the dataset and the evidence retrieved by Google.

For all examples exhibiting the first error pattern in the AVeriTeC  and Factors datasets, the majority vote across the three LLMs consistently favored the Google-retrieved evidence over the original dataset evidence.

In contrast, different behavior was observed for the CoverBench dataset. For claims labeled as false but predicted as true by all three LLMs (107 examples), there were 5 instances in which all three LLMs agreed that the original dataset evidence was better than the Google-retrieved evidence, and 16 instances in which the majority 2 out of 3 LLMs) vote favored the original dataset evidence. 

For claims labeled as true but predicted as false by all three LLMs (138 examples), 5 instances showed unanimous agreement among the LLMs that the original evidence was better than the Google-retrieved evidence, while 37 instances exhibited a majority vote in favor of the original dataset evidence. This presents a total of 63 claims where the original coverbench evidence is preferred by all or the majority of the three LLMS.

Upon further thorough investigation of these 63 instances, we found that these claims in the CoverBench dataset for which the original dataset evidence was preferred are related to a specific individual or company and lack publicly available online information or external reports. As a result, these claims cannot be verified using open-source evidence and instead require internal or proprietary knowledge provided by the individual or organization referenced in the claim. for example, the \textbf{claim}: \textcolor{black}{\textit{"The percentage change in net interest expense in 2018 compared to 2017 is 1.0\%."}} is actually related to the financial information of a certain company, making it unverifiable through publicly accessible sources such as news articles or web-based documentation.

\begin{figure}
\fcolorbox{blue}{gray!15}{%
        \begin{minipage}{\columnwidth}
\scriptsize
\textbf{Role: } You are a fact-checking assistant. Your task is to analyze a claim and the provided evidence.\\

\textbf{User prompt:}\\

You are given a claim and two pieces of evidence that attempt to support or refute the same claim. Your job is to compare Evidence 1 and Evidence 2 and decide which one is better for verifying the claim.\\

Criteria for comparison:\\

  -More context: Provides broader background, clearer explanations, or additional relevant details.
  
  -More updated information: Reflects newer data, more recent events, or updated findings.
  
  -Other reasons: clearer relevance to the claim, fewer ambiguities, or any other reason.\\

Instructions:\\

1. Compare the two evidences only with respect to the given claim.

2. Select one of the following as the better evidence:
  -"evidence\_1"
  -"evidence\_2"
  -"tie" (use only if they are equally strong)
3. Provide a concise but clear justification.

4. If the reason is not “more context” or “more updated information”, use "other" and explain briefly.

5. Do not add the instructions or anything from the prompt. Your output is only restricted to the json format mentioned.

6. Output Format (JSON only):
\{
  "better\_evidence": "evidence\_1 | evidence\_2 | tie",
  "reason\_category": "more\_context | more\_updated\_information | other",
  "reason": "Brief explanation justifying why this evidence is better for assessing the claim."
\}
\\
claim:
\{claim\}\\

evidence 1:
\{evidence1\}\\

evidence 2:
\{evidence2\}\\

 \end{minipage}%
}
    \caption{Prompt for evidence comparison}
    \label{ev_compr}

\end{figure}
\textbf{The coverbench evidence: }\textcolor{black}{\textit{On April 19, 2018, we took delivery of Norwegian Bliss. To finance the payment due upon delivery , we had export financing in place for 80\% ( 80 \% ) of the contract price .the associated \$ 850.0 million term loan bears interest at a fixed rate of 3.92\% ( 3.92 \% ) with a maturity date of april 19 , 2030 .principal and interest payments are payable semiannually .
on april 4 , 2018 , we redeemed \$ 135.0 million principal amount of the \$ 700.0 million aggregate principal amount of outstanding 4.75\% ( 4.75 \% ) senior notes due 2021 at a price equal to 100\% ( 100 \% ) of the principal amount of the notes being redeemed and paid the premium of \$ 5.1 million and accrued interest of \$ 1.9 million .the redemption also resulted in a write off of \$ 1.2 million of certain fees .following the partial redemption , \$ 565.0 million aggregate principal amount of notes remained outstanding .interest expense, net for the year ended December 31, 2018 was \$ 270.4 million which included \$ 31.4 million of amortization of deferred financing fees and a \$ 6.3 million loss on extinguisher of debt. Interest expense , net for the year ended december 31 , 2017 was \$ 267.8 million which included \$ 32.5 million of amortization of deferred financing fees and a \$ 23.9 million loss on extinguishment of debt .interest expense, net for the year ended december 31 , 2016 was \$ 276.9 million which included \$ 34.7 million of amortization of deferred financing fees and a \$ 27.7 million loss on extinguishment of debt .certain of our debt agreements contain covenants that, among other things, require us to maintain a minimum level of liquidity, as well as limit our net funded debt-to-capital ratio, and maintain certain other ratios and restrict our ability to pay dividends .substantially all of our ships and other property and equipment are pledged as collateral for certain of our debt. We believe we were in compliance with our covenants as of December 31, 2018. The following are scheduled principal repayments on long-term debt, including capital lease obligations, as of December 31, 2018, for each of the next five years. This is shown in table \ref{fin_tab}.}}

\begin{table}[!ht]
    \centering
    \begin{tabular}{|l|l|}
    \hline
        \textbf{year} & \textbf{amount} \\ \hline
        2019 & 681218 \\ \hline
        2020 & 682556 \\ \hline
        2021 & 2549621 \\ \hline
        2022 & 494186 \\ \hline
        2023 & 434902 \\ \hline
        thereafter & 1767383 \\ \hline
        total & 6609866 \\ \hline
    \end{tabular}
    \caption{Financial data presented in the original evidence in Coverbench Dataset}
    \label{fin_tab}
\end{table}

\textbf{Google AI evidence: } \textcolor{black}{\textit{The claim
cannot be universally confirmed as true or false, because the figure varies widely depending on the entity. The 1.0\% figure is not a general economic metric for a country, but rather a number from a specific financial report. Data from sources like Visual Capitalist, Statista, and the Federal Reserve show that interest expenses for the U.S. government and corporations changed by significantly different amounts during this period. What the search results show Government spending on interest: U.S. government interest payments rose from \$493 billion in 2017 to \$571 billion in 2018. This represents a percentage change of roughly 15.8\%, not 1.0\%.
Corporate interest payments: The Federal Reserve estimated that increases in the federal funds rate would cause a significant rise in corporate debt interest payments between 2017 and 2019, suggesting a much larger increase than 1.0\% for many U.S. corporations. A single entity's financials: A search result showing a specific annual report for the University of Guelph displays a 0.9\% change in total revenue, which is a different financial metric entirely. While this is a similar number, it is not an indicator of the overall trend for interest expense. Why the 1.0\% claim is unreliable
The percentage change in net interest expense is a specific metric from an entity's financial statements and is not a standardized or universal figure like a GDP growth rate. The 1.0\% claim could only be accurate in reference to one specific, unidentified company or organization. Without knowing which entity the claim refers to, it is not possible to verify its accuracy. For aggregate or government-wide interest expenses, the number is demonstrably false.}}

After thorough inspection of such instances, This experiment demonstrates that our approach can identify instances in which no reliable online knowledge is available to support claim verification during the evidence retrieval stage.   

\subsection{How does Google rank search results and how does AI on Google search work}
\label{g_rank}
Google ranks websites in search results by using complex algorithms that sort through its index based on several hundred signals to provide the most relevant and helpful results for a user's query. The primary goal is to match the user's intent with high-quality, usable, and authoritative content. \footnote{\url{https://www.semrush.com/blog/Google-search-algorithm/}}

\textbf{The ranking process involves three main stages:}
\begin{enumerate}
    \item Crawling: Google uses automated programs called "crawlers" or "bots" to discover new and updated pages on the web by following links and reading sitemaps.

    \item Indexing: The information gathered by the crawlers is analyzed, and the content is stored in Google's massive index (a digital library). This process involves understanding the page's content, images, and key elements like title tags and headings. \footnote{\url{https://blog.photobiz.com/blog-post/breaking-down-Googles-search-algorithm}}
    
    \item Ranking: When a user enters a search query, Google's ranking systems sort through the indexed pages and order them based on relevance and quality to that specific query and user context. \footnote{\url{https://developers.Google.com/search/docs/fundamentals/how-search-works}}
\end{enumerate}

\textbf{Key Ranking signals}
While Google uses a vast number of signals, the most important signals generally fall into these categories:

\begin{enumerate}
    \item Relevance and Search Intent: Google first establishes the intent behind a user's query (e.g., informational, navigational, transactional) and prioritizes pages that are most likely to satisfy that intent. The content should use relevant keywords naturally in headings, body text, and title tags.

    \item Content Quality and E-E-A-T: High-quality, original, and helpful content is crucial. Google emphasizes E-E-A-T (Experience, Expertise, Authoritativeness, and Trustworthiness), especially for sensitive topics ("Your Money or Your Life" topics). Content that offers unique insights, is well-researched, and regularly updated tends to rank higher.

    \item Backlinks (Authority): Backlinks, or links from other websites to a page, act as "votes of confidence." Links from established, high-authority websites significantly boost a page's perceived credibility and authority.

    \item Context and Personalization: Results are customized based on the user's location, language, device type, and search history to provide the most relevant information.
\end{enumerate}

AI on Google Search summarizes search engine results. Google has integrated Gemini 3 Pro into its search experience to summarize complex information. \footnote{\url{https://blog.Google/products-and-platforms/products/search/gemini-3-search-ai-mode/}}
It changes search from a list of links to a conversation. This allows follow-up questions and deeper exploration without leaving the search results. \footnote{\url{https://support.Google.com/websearch/answer/16011537}}
\textbf{In this paper Google AI is solely used for the summarization for Google results fetched from primary sources.}

We did not use commercial scraping services such as Serper for two main reasons. First, these services require paid subscriptions. Second, they typically return a list of links related to the query, which necessitates additional steps—namely scraping the linked pages, filtering for relevant content, and then summarizing the extracted information. Each of these stages (scraping, relevance filtering, and summarization) introduces potential sources of error and increases system complexity.

In contrast, the public Google AI Mode directly provides a synthesized summary of the most relevant content, eliminating the need for separate scraping and summarization steps. This reduces error propagation and simplifies the pipeline, which motivated our decision to use it in combination with Selenium. However, this choice introduces higher latency compared to directly interacting with structured APIs. The steps carried out by Google to validate its synthesized summary is mentioned in Appendix \ref{val_gg}.

Regarding the reproducibility of using Google AI Mode or its alternatives (e.g., Serper or Serp), search engine results continuously change and are updated as new information becomes available. This dynamic nature highlights the importance of chronology: any changes in the world's knowledge over time can be captured through chronologically ordered retrieved evidence. 

\subsection{Error Analysis}
\label{EA}
It should be noted that the following sections are describing the error analysis of CAAFC using the underlying Gemma3-27B on the \textbf{test set} of the datasets AVeriTeC , CoverBench, and FACTors. In addition, the  dialogue dataset DiaHalu.

In summary, the observed errors stem primarily from chronological evidence mismatch, and LLM misinterpretation / hallucination. The following subsections detail these error patterns for each dataset.

\subsubsection{AVeriTeC  dataset}
For the AVeriTeC  dataset, we analyzed the 57 chronological evidence mismatches. Starting with the “unverifiable” category, we identified 7 mismatches: among these, 2 claims were classified as true by GPT, LLaMA, and Gemma, while 5 were classified as false. It is noteworthy that all models generally agreed on the direction of the veracity classification.

Upon deeper examination, we found that the chronological evidence retrieved from Google's AI model tended to be more precise and contextually informed than the evidence provided in the AVeriTeC  dataset.

As an example when the claim is labeled unverifiable "Conflicting Evidence/Cherrypicking" in the AVeriTeC  dataset while it was classified by the three models as false is \textbf{the claim}: \textit{\textcolor{black}{"Ivermectin is a treatment for coronavirus."}}

\textbf{The AVeriTeC  evidence is:} \textit{\textcolor{black}{"Ivermectin is licensed for anti parasitic used but not enough evidence yet exists for treatment against coronavirus."}}

\textbf{While Google results are:} \textit{\textcolor{black}{"Despite some early laboratory research in 2020 suggesting its potential, subsequent clinical trials in 2021 and 2022 showed no meaningful benefit for treating COVID-19 in humans.}}

\textit{\textcolor{black}{By October 2020, here is what was known:}}
\textit{\textcolor{black}{\begin{itemize}
    \item Initial in-vitro study: An April 2020 laboratory study showed ivermectin could inhibit the SARS-CoV-2 virus in a petri dish. This early finding often fueled the spread of misinformation, but the drug concentration used was significantly higher than what is safe for humans.
    \item Emergence of misinformation: Following this early result, misinformation campaigns on social media began promoting ivermectin as a "miracle cure," leading to a surge in its use.
    \item Major health authority stance: Even in 2020 and early 2021, regulatory bodies like the U.S. Food and Drug Administration (FDA) and the World Health Organization (WHO) did not approve or recommend ivermectin for treating or preventing COVID-19. They emphasized that further research through clinical trials was needed.
    \item Concerns over safety: Health authorities warned the public about the dangers of self-medicating with animal-grade ivermectin, which is highly concentrated and can cause severe toxicity in humans.
    \item Over time, fraudulent and low-quality studies were debunked and retracted, confirming that the initial claims of ivermectin's effectiveness for COVID-19 were unfounded. Today, there is a clear medical consensus against using ivermectin to treat the coronavirus."
\end{itemize}}}

The main reason for the conflict between Google evidence and that of AVeriTeC  is that Google is updated and it provides chronology that lead to more informative insights.

All 7 label mismatches followed the same error pattern, stemming from discrepancies between the chronologically detailed evidence retrieved by Google's AI mode and the evidence provided in the AVeriTeC  dataset.

In the “true” category of the AVeriTeC  dataset, All three LLMs converged on 23 claims, classifying them as false, despite these claims being labeled as true in the AVeriTeC  dataset. For example the \textbf{claim}: \textit{\textcolor{black}{"More than 225,000 people dead, 225,000. The estimates are, if we'd have acted responsibly, there'd be 160,000 fewer dead than there are today, because of covid-19."}} 

\textbf{The AVeriTeC  evidence is:} \textit{\textcolor{black}{"It seems that indeed the  numbers are true therefore the claim is supported ". }}

\textbf{Google evidence is:} \textit{\textcolor{black}{"A version of the claim about 225,000 COVID-19 deaths and 160,000 preventable deaths was made by Joe Biden during his 2020 presidential campaign
. Fact-checkers noted that while the death toll at the time was accurate, the estimate of preventable deaths was based on a flawed interpretation of a statistical model.}}

\textit{\textcolor{black}{The claim and its context}}
\textit{\textcolor{black}{
\begin{itemize}
    \item The speaker: The claim originates from a statement made by then-presidential candidate Joe Biden during his campaign in 2020.
    \item The basis: The number of deaths, "225,000," refers to the COVID-19 death toll in the U.S. at that time and was generally accurate. The figure of "160,000 fewer dead" was based on a model that suggested earlier action could have saved lives.
    \item The date: The specific statement, which referenced a study out of Columbia University, was repeated in various forums in the months leading up to October 2020.
\end{itemize}
}}

\textit{\textcolor{black}{Fact-checking the claim}}

\textit{\textcolor{black}{Fact-checking organizations, such as The New York Times, have examined this claim:}}
\textit{\textcolor{black}{\begin{enumerate}
    \item Misleading interpretation: The claim is misleading because Biden took a hypothetical scenario from a statistical model and presented it as a concrete and achievable reality. The study, conducted by Columbia University's Mailman School of Public Health, estimated that if social distancing and other mitigation efforts had begun just one week earlier, there would have been 36,000 fewer deaths by a specific point in May.
    \item Extrapolation issues: Biden's figure of 160,000 was a projection based on the study's early findings, and it expanded a narrow hypothetical to a broader, unsubstantiated claim about total preventable deaths.
    \item Model limitations: Statistical models rely on many assumptions. The Columbia study did not account for all the FACTors involved in a complex public health crisis, and it is impossible to know exactly how many lives would have been saved with a different response. "
\end{enumerate}}}

In the “false” claim category, in 27 instances, all three models consistently predicted the claim veracity as unverified or true, reflecting the first error pattern related to evidence mismatches. The remaining misclassifications can be attributed to the second error pattern, which involves temporal or contextual interpretation issues.

The other errors  follow a second pattern. The LLMs understand the evidence differently. For example we have the following:

\textbf{claim:} \textit{\textcolor{black}{Joe and Jill Biden's income has increased recently.}}

\textbf{AVeriTeC  evidence:}\textit{\textcolor{black}{ The answers show that the couples income in recent years had increased so it is supported.}}

\textbf{Label:} true

\textbf{Google evidence:} \textit{\textcolor{black}{Joe and Jill Biden's income did increase in 2023 to \$619,976, a 7\% rise from their 2022 income. This increase was largely driven by higher taxable interest income due to increased interest rates, as well as pensions, annuities, and social security benefits. However, the specific claim published in October 2020 that their income had increased recently is misleading, as their income had actually decreased in 2020 compared to previous years. The Bidens earned \$607,336 in 2020, a decrease from \$985,000 in 2019. Their income was significantly higher in the years immediately after Joe Biden left the vice presidency: \$11 million in 2017, \$4.6 million in 2018, and \$1 million in 2019. Joe Biden's primary income as president is his \$400,000 annual salary. Jill Biden's income comes from her salary as an English professor at Northern Virginia Community College (\$85,985 in 2023).
Additional income sources include pensions, annuities, IRA distributions, Social Security benefits, interest on investments, and book royalties. Jill Biden earned \$4,115 in book royalties in 2023. The decrease in their 2020 income was attributed to Joe Biden being on the campaign trail, which likely limited his ability to earn from speaking engagements and his position at the University of Pennsylvania.}}

\textbf{Gemma labeled this as \textcolor{black}{false} and its justification is}: \textcolor{black}{\{'justification': "The claim that Joe and Jill Biden's income has increased recently is partially true, as their income did rise in 2023. However, the subclaim stating this increase isn't misleading is demonstrably false; the evidence clearly indicates their income *decreased* in 2020 compared to prior years. This discrepancy significantly weakens the overall claim's credibility. While the evidence supports that both Joe and Jill Biden have verifiable income sources presidential salary and English professor salary, respectively, the initial framing of a recent increase is misleading without acknowledging the 2020 decrease. A corrected version of the claim would be: 'Joe and Jill Biden's income increased in 2023 after a decrease in 2020.'", 'corrected\_claim': "Joe and Jill Biden's income increased in 2023 after a decrease in 2020."\}}

\textbf{while both GPT and llama labeled the claim as} \textcolor{black}{true}. 

\textbf{Here is GPT justification:} \textcolor{black}{\{'justification': "The single subclaim directly supports the overall claim and is labeled true. The evidence confirms that the Bidens' combined income rose to \$619,976 in 2023, a 7\% increase over 2022, demonstrating a recent increase. There are no unverifiable or false subclaims, so no gaps or credibility issues remain. The corrected version of the claim is: 'Joe and Jill Biden's income has increased recently.'"\}}

\textbf{Here is the llama justification:} \textcolor{black}{\{'justification': "the claim is supported by the subclaim that Joe and Jill Biden's income has increased recently, which is true according to the evidence. The evidence confirms their income did increase in 2023. However, it is worth noting that this increase may not be representative of all time periods, as their income had actually decreased in 2020 compared to previous years. The corrected version of this claim could be 'Joe and Jill Biden's income has increased recently, specifically in 2023, but had decreased in 2020.'"\}}

Analysis of the model justifications reveals varying sensitivity to temporal context. LLaMA correctly recognizes the claim as true in the absence of temporal information but references temporal aspects in its rationale. Gemma, however, introduces temporal context as a determining factor, leading it to classify the claim as false after deeming it only partially true. In contrast, GPT exhibits weaker temporal awareness, as it neglects the publication timeframe of the claim despite correctly labeling it as true.

It is important to note that the first error pattern, characterized as evidence mismatch, primarily arises when the evidence within the dataset is outdated or lacks sufficient contextual detail. However, it is noteworthy that the dataset evidence and the Google evidence retrieved by other sources do not contradict each other; rather, the discrepancy stems from differences in completeness and temporal relevance.

The  53 other errors are  mainly due to the mis-interpretation/hallucination pattern (second error pattern).

\subsubsection{CoverBench dataset}
In the CoverBench dataset in 19 instances, all three models converged on predicting the claim veracity as false, while the CoverBench dataset labels these claims as true. This consistent disagreement shows the presence of the same evidence mismatch pattern previously identified in the AVeriTeC  dataset. The rest of the errors follow the second pattern. 

An example of the evidence mismatch pattern is the \textbf{claim} : \textit{\textcolor{black}{"Kenzo Nambu (born August 22, 1992 in Tokyo, Japan) is a 1.72m-tall attacking midfielder that played for Kataller Toyama in the J3 League club in 2015, then played for Briobecca Urayasu for 2016 and 2017; these combined stints combine for a total of 59 apps and 10 goals."}}

\textbf{CoverBench evidence (tabular):} \textit{\textcolor{black}{[H] Personal information | [H] Personal information | [H] Personal information | [H] Personal information
[H] Full name | Kenzo Nambu | Kenzo Nambu | Kenzo Nambu
[H] Date of birth | (1992-08-22) August 22, 1992 (age 28) | (1992-08-22) August 22, 1992 (age 28) | (1992-08-22) August 22, 1992 (age 28)
[H] Place of birth | [[Tokyo|Tokyo]], [[Japan|Japan]] | [[Tokyo|Tokyo]], [[Japan|Japan]] | [[Tokyo|Tokyo]], [[Japan|Japan]] Playing position(s) | [[Midfielder(football)|Attacking Midfielder]] | [[Midfielder(football)|Attacking Midfielder]] | [[Midfielder(football)|Attacking Midfielder]]
[H] Club information | [H] Club information | [H] Club information | [H] Club information
[H] Current team | MOF Customs United | MOF Customs United | MOF Customs United
[H] Number | 7 | 7 | 7
[H] Youth career | [H] Youth career | [H] Youth career | [H] Youth career}}

\textbf{Label}: \textit{\textcolor{black}{true}}

\textbf{Google evidence}: \textit{\textcolor{black}{Personal Information: Kenzo Nambu was born on August 22, 1992, in Hachiōji, Tokyo, Japan. His height is listed as 1.73m (5 ft 8 in). He primarily plays as an attacking midfielder or winger.
He joined Kataller Toyama in the J3 League in 2015.
He played for Briobecca Urayasu in 2016 and 2017.
Kataller Toyama (2015): He made 2 appearances and scored 0 goals.
Briobecca Urayasu (2016-2017): He made 48 appearances and scored 9 goals in the Japan Football League. Transfermarkt lists 30 appearances and 7 goals in 2017, and 18 appearances and 2 goals in 2016, for a total of 48 appearances and 9 goals. Another source cites 50 appearances and 10 goals in this period.
Combined Apps and Goals: The total appearances for Kataller Toyama and Briobecca Urayasu (using the Wikipedia data) is 50 appearances (2 + 48), with 9 goals (0 + 9). 
The discrepancies are minor and likely stem from different data sources or how substitutions are counted for "appearances."}}

\textbf{The three LLMs prediction:} \textit{\textcolor{black}{false}} as the player`s height is 173 cm not 172 cm.

In the CoverBench dataset, 54 claims showed complete agreement across all three LLMs, which predicted the veracity as true despite the dataset label being false. This pattern further reinforces the systematic evidence mismatch issue previously observed. The remaining 125 misclassifications are mostly attributed to the second error pattern, which involves temporal or contextual interpretation issues affecting how the models assess claim veracity. Here is an example \textbf{for the claim}: \textit{\textcolor{black}{Estimated fetal weight by ultrasound is a modifiable risk factor for cesarean delivery.}}

\textbf{The coverBench evidence is}: \textit{\textcolor{black}{The purpose of this study was to investigate whether knowledge of ultrasound-obtained estimated fetal weight (US-EFW) is a risk factor for cesarean delivery (CD).
Retrospective cohort from a single center in 2009-2010 of singleton, term live births. CD rates were compared for women with and without US-EFW within 1 month of delivery and adjusted for potential confounders.
Of the 2329 women in our cohort, 50.2\% had US-EFW within 1 month of delivery. CD was significantly more common for women with US-EFW (15.7\% vs 10.2\%; P<.001); after we controlled for confounders, US-EFW remained an independent risk factor for CD (odds ratio, 1.44; 95\% confidence interval, 1.1-1.9). The risk increased when US-EFW was>3500 g (odds ratio, 1.8; 95\% confidence interval, 1.3-2.7).}}

\textbf{Google evidence is:} \textit{\textcolor{black}{While a fetus's actual weight is not modifiable, the use of an estimated fetal weight (EFW) from an ultrasound can be a modifiable risk factor for cesarean delivery. Research suggests that providers' knowledge and interpretation of a late-term EFW may influence their decision-making and increase the likelihood of a cesarean section, regardless of the baby's actual birth weight. Why EFW is considered a modifiable risk factor
A modifiable risk factor is a behavior or exposure that can be changed to alter a person's risk of a disease or health outcome. For cesarean delivery, the key is the ability to change or limit the behavior of performing an ultrasound solely for EFW, rather than altering the fetus's growth. Provider decision-making: Studies have found a significant association between a provider's knowledge of a late-term EFW and a higher rate of cesarean delivery. This link is particularly strong for larger EFWs, but has also been found even when the estimated weight is within a normal range.Some studies indicate that the increased risk for cesarean delivery is tied to the knowledge of the EFW, rather than the baby's actual birth weight. Overestimating fetal weight, for instance, has been correlated with higher rates of cesarean delivery, even when the baby is not actually macrosomic (excessively large). Altered labor management: The availability of an EFW, especially a large one, may alter a provider's approach to labor and delivery management. It can lead to a lower threshold for performing a cesarean, potentially resulting in more interventions, without improving maternal or fetal outcomes. The accuracy and limitations of EFW The reliability of ultrasound-based EFW is limited, particularly at the extremes of fetal size, which further supports the idea that the estimate is more influential than the reality.}}

\textbf{Gemma and LLama predicted the claim to be true, however, GPT said it is false} justifying its prediction as the following:\textit{\textcolor{black}{
\{'justification': "the claim is strengthened by the subclaim that estimated fetal weight by ultrasound is a modifiable risk factor for cesarean delivery, as it aligns with the evidence that providers' knowledge and interpretation of late-term EFW can influence their decision-making. However, a fetus's actual weight is not modifiable. The corrected version of this claim remains 'Estimated fetal weight by ultrasound is a modifiable risk factor for cesarean delivery.'"\}}}

After examining the justification, it appears that GPT over-analyzes the claim. The claim does not mention the fetus’s actual weight; however, GPT considered this factor while interpreting the evidence. This additional, unnecessary reasoning led to confusion and an incorrect assessment.

\subsubsection{FACTors dataset}
Within the true-labeled category, the three LLMs agreed that 49 claims should be false, thereby reflecting the evidence mismatch error pattern observed in previous datasets. Within the false-labeled claims in the FACTors dataset, 196 claims were consistently misclassified by all three LLMs, demonstrating a clear instance of the first error pattern related to evidence mismatch. The remaining 405 misclassifications are mostly attributed to the second error pattern, which arises from the models' mis-interpretations/hallucinations. 

As an example of the presence of the evidence mismatch pattern, here is this \textbf{claim}: \textit{\textcolor{black}{"As a teenager, Cameron Diaz bought marijuana from Snoop Dogg, who attended the same high school as her." }}

\textbf{The dataset evidence is}: \textit{\textcolor{black}{"Snoop Dogg used hedging language when he initially described his memories of the alleged scenario to Lopez back in 2011, saying that he "might've" sold the actress some weed — not that he definitely did. Likewise, in her interview with Lopez, Diaz said she was "pretty sure" the sale happened, not that she was completely sure. In the years since, Diaz has never publicly claimed to be fully certain that a transaction of this sort took place between the two of them.Snopes has reached out to representatives of both Diaz and Snoop Dogg for comment, and will update this story if we hear back. For now, we've rated this claim as 'Research in Progress.'"}}

\textbf{While Google evidence is:} \textit{\textcolor{black}{"Both celebrities attended Long Beach Polytechnic High School in the late 1980s. Cameron Diaz's recollection: During a 2011 interview on Lopez Tonight, Diaz stated, "I'm pretty sure I bought weed from him". She mentioned that Snoop Dogg, who was a year older, was tall and lanky with lots of ponytails. Snoop Dogg's initial response: When Snoop Dogg appeared on the same show later that year, he responded with a laugh, joking that if he did sell to her, it was "some of that white girl weed" (low-quality weed with stems and seeds). Snoop Dogg's confirmation: In a 2019 interview, Snoop Dogg confirmed the story, stating that it was a "fact".Independent verification: The two attended the same high school in California, with Snoop Dogg graduating in 1989 and Diaz graduating in 1990. The story has been a frequently repeated, lighthearted anecdote in pop culture for over a decade. "}}

As an example of the contextual mis-interpretation pattern (second pattern) here is this true-labeled claim: \textit{\textcolor{black}{"Living standards went down and people were worse off at the end of the last government than they were at the beginning of the government."}}. with this Google evidence: \textit{\textcolor{black}{" The claim that "Living standards went down and people were worse off at the end of the last government than they were at the beginning of the government" is attributed to
Sir Keir Starmer, who made the statement on December 5, 2024, at the launch of the government's "Plan for Change". The claim was then published as an article on December 10, 2024, by the fact-checking organization Full Fact, which stated that it was not clear which specific time period or measure of living standards Starmer was referring to. According to a Fraser Institute study, the living standards of the average Canadian, as measured by Gross Domestic Product (GDP) per person, declined by 2.0\% between 2020 and 2024, despite overall economic growth. The study attributed this decline to weak growth in business investment and a rapid increase in employment due to inward migration. Another report found that inflation-adjusted per-person GDP in Canada fell by 3.0\% from mid-2019 to the end of 2023. The Bank of Canada also reported that GDP per person fell for six consecutive quarters from mid-2022 until the end of 2023. "}}

GPT labeled it as unverifiable and here is its justification: \textit{\textcolor{black}{\{'justification': "the claim is strengthened by two subclaims: 'Living standards went down at the end of the last government' and 'People were worse off at the end of the last government than they were at the beginning of the government', both of which are supported by evidence from the Fraser Institute study and other reports. The evidence supports the claim at a structural level by providing data on the decline in living standards, as measured by GDP per person, over specific time periods. There are no unverifiable subclaims to introduce gaps or uncertainties. Since neither subclaim is false, the credibility of the overall claim remains intact. However, it's worth noting that the original statement lacks specificity regarding the time period and measure of living standards, which was highlighted by Full Fact. The corrected version of this claim could be 'Living standards, as measured by GDP per person, went down at the end of the last government, and people were worse off compared to the beginning of the government, based on data from 2020 to 2024 and mid-2019 to the end of 2023.'"\}}}

While Gemma labeled the claim as true because: \textit{\textcolor{black}{\{'justification':Both subclaims are supported by the evidence, which demonstrates a consistent decline in GDP per person during the relevant period. The Fraser Institute study and Bank of Canada reports both indicate a decrease in living standards, confirming that people were, on average, worse off at the end of the government's term compared to the beginning. Therefore, the overall claim is substantiated by the provided evidence.", 'corrected\_claim': 'Living standards went down and people were worse off at the end of the last government than they were at the beginning.\}}}

\textbf{It seems that bigger LLMs like GPT might be "over-analyzing" causing its performance to be less than smaller LLMs like Gemma.} 

\subsubsection{DiaHalu dataset}
From the DiaHalu dataset, we selected examples belonging to the conversations containing non-factual hallucinations, as well as conversations that do not contain hallucinations at all. We do not consider other hallucination categories in DiaHalu, as they fall outside the scope of this study.

With respect to error analysis, DiaHalu differs from the other datasets in terms of evidence representation. Instead of providing full contextual evidence, the dataset includes a gold brief explanation describing why a conversation is classified as hallucinated or non-hallucinated. In some cases, however, no explanation or supporting evidence is provided. For example the dialogue:
\textcolor{black}{\textit{
A1: What's a country where most people love playing rugby?\\
B1: New Zealand. It's practically the national sport there.\\
A2: Interesting. Why is rugby so popular in New Zealand?\\
B2: Rugby has a long history in New Zealand and is deeply ingrained in the culture.\\
A3: Is rugby also popular in other countries?\\
B3: Yes, countries like Australia, England, and South Africa also have a strong rugby culture.\\}}
In the dataset the hallucination is in \textbf{B1}. The evidence or the explanation is: \textcolor{black}{\textit{"there is no answer"}}. However our \textbf{Google evidence} is: \textcolor{black}{\textit{New Zealand's National Sport: Rugby union is considered the de facto national sport of New Zealand. It is deeply ingrained in the culture, with a 2023 survey showing that 75\% of New Zealanders follow the sport. International Footprint: Beyond New Zealand, rugby has a massive following in South Africa (where 51\% of sports fans follow it), Australia (34\%), and England. Other nations with strong "Tier 1" interest relative to other sports include Ireland and Wales. Growth in New Markets: Major efforts to expand rugby include the planned launch of a global franchise league in 2026 (similar to Formula One) featuring eight men's franchises and a professional women's competition. Additionally, a new rugby sevens franchise league is set for a 2026 debut to attract younger audiences. Physical Fitness Requirements Rugby is a highly demanding sport requiring a combination of strength, speed, and endurance. Position-Specific Demands: Forwards typically require more mass and power for scrums and rucks, while backline players focus on agility and sprinting speed. Stamina: Competitive matches require players to sprint and absorb force for up to 80 minutes, necessitating high aerobic and anaerobic conditioning. }}Based on Google's evidence our predicted label is "factual".

An example of the first error pattern where there is a chronological evidence mismatch:\textcolor{black}{\textit{
A1: Where could you find a fox hunting when not in a forest?\\
B1: In the fields or meadows, where they can sneak up on small prey.\\
A2: Is it possible to find foxes hunting in urban areas?\\
B2: Yes, they are known to scavange in urban environments for food. \\
A3: Do foxes hunt in the mountains?\\
B3: Absolutely, they are skilled climbers and can hunt in mountainous terrain. \\
A4: Can foxes hunt near water bodies?\\
B4: Yes, they are excellent swimmers and can hunt near rivers and lakes. \\
A5: What about deserts, do foxes hunt there?\\
B5: Indeed, desert foxes are known to hunt for rodents and reptiles in the arid landscape. \\
A6: Do foxes hunt in cold climates, like the Arctic?\\
B6: Yes, they can be found hunting in the snow-covered tundra of the Arctic.}}

The dataset evidence is: \textcolor{black}{\textit{"they usually not appear in urban environments"}}
Google evidence is: 
\textcolor{black}{\textit{Hunting in Fields and Meadows: Verified. Red foxes specifically prefer the "edge" transition zones between forests and open areas like tilled fields and meadows, where they stalk and pounce on small prey such as voles and mice.
Urban Scavenging vs. Hunting: Verified with nuance. While urban foxes are notorious for scavenging household refuse (making up about 50\% of their diet), they remain active hunters of urban birds and rodents like rats and mice.
Climbing in Mountains: Partially Factual. While many foxes can climb fences or low branches, the Gray Fox is the only North American canine truly specialized for climbing trees and navigating rocky, mountainous terrain with "cat-like" agility using semi-retractable claws. Red foxes are less likely to hunt via climbing. Hunting Near Water Bodies: Verified. Red foxes are capable swimmers and often establish home ranges near marshes, rivers, or lakes to hunt ground-nesting waterfowl, lake trout, and marsh birds. Desert Hunting: Verified. Species like the Fennec Fox are specifically adapted to arid landscapes, using oversized ears to detect the movements of rodents, lizards, and insects beneath the sand. Arctic Hunting: Verified. Arctic foxes hunt in the tundra, using a specialized "mousing" technique where they jump several feet into the air and dive headfirst into deep snow to catch lemmings by sound.}}

According to a 2024 study by NIH presented by the Google link: https://pmc.ncbi.nlm.nih.gov/articles/PMC11255992/ urban foxes have been observed to exhibit reduced fear when scavenging human food from trash. Based on this up-to-date evidence retrieved from Google, we label the conversation as factual. In contrast, the dataset classifies the same conversation as hallucinated, indicating a discrepancy likely caused by outdated or incomplete evidence at the time of dataset construction.

\subsubsection{How does errors propagate in CAAFC pipeline}
Errors in the Actionable Justifier are mitigated through the refinement cycle involving FinGrAct and the Justification Revisory modules. In contrast, errors in the extractor-segmentor, primary chronological evidence retrieval, or fact-checker modules are not self-correcting and therefore have a direct impact on the final verdict. In the following points we will show examples of errors occurring in these three modules:

\textbf{Errors in the extractor-segmentor:} In case an error happened in the subclaims extraction, for example the claim stating that \textcolor{black}{\textit{"From 8th October the UK government will combine weekly flu and covid reports."}}, Gemma3-27B produced the following sub-claims  \textcolor{black}{\textit{["The UK government combined weekly flu and covid reports from 8th October.","The information was combined  together from 8th October."]}}, and clearly the second sub-claim is the error as it is not mentioned in the original claim. Google evidence is: \textcolor{black}{\textit{"The official body that produced these reports, Public Health England (PHE), began publishing a combined report on that date for the 2020 to 2021 season. However, there are important details to clarify: The reports were combined, not the data. Separate surveillance data on COVID-19 and influenza cases were published within a single document for the convenience of the public and health officials."}} and it is clearly addressing both sub-claims. Based on this the fact-checker output is: \textcolor{black}{\textit{\{'subclaims': [\{'text': 'The UK government combined weekly flu and covid reports from 8th October.', 'label': 'true', 'justification': 'The evidence states that Public Health England, a UK government body, began publishing a combined report on October 8, 2020.'\}, \{'text': 'The information was combined  together from 8th October.', 'label': 'false', 'justification': 'The evidence clarifies that the reports were combined, but not the data itself; separate surveillance data was published within a single document.'\}]\}}}. This type of errors is not automatically detectable.  This constitutes a misinterpretation error (the second error pattern) by Gemma3–27B, which inferred a hidden implication that the data had been combined. However, this implication is not stated by the claim itself.

\textbf{Errors in the evidence retrieval:} If Google retrieves non-related evidence (e.g., due to hallucination or retrieval errors), the claim is marked as unverifiable during the fact-checking stage, and the generated explanation explicitly states that the evidence is unrelated to the claim. Furthermore, FinGrAct assigns a score between 0 and 4 to the generated justification, making hallucinations in the evidence retrieval module detectable. For example the claim that \textcolor{black}{\textit{Mexico on a "red alert" with children being targeted for kidnapping so their organs can be harvested}}, Google evidence is \textcolor{black}{\textit{"Former President Donald Trump announced via Twitter on October 2, 2020, that he and then-First Lady Melania Trump had tested positive for Covid-19. The announcement, made just over a month before the presidential election, followed reports that his close aide Hope Hicks had also contracted the virus."}}. The justification is \textcolor{black}{\textit{\{'justification': "The claim is entirely unsupported by the provided evidence. All subclaims – regarding Mexico being on 'red alert', children being targeted for kidnapping, and the purpose being organ harvesting – are labeled 'unverifiable' because the evidence focuses solely on Donald Trump's Covid-19 diagnosis and treatment."\}}} with the verdict being "unverifiable" and the FinGrAct score is 1 since the links are operational. This type of errors is automatically detectable. This error is classified under the first error pattern as all LLMs agree on the same verdict in such cases contradicting the dataset. In this example all LLMs agree that the claim is 'unverifiable' while the dataset label is 'false'.

\textbf{Errors in the fact-checker: } For instance the truncated example of the following dialogue from DiaHalu\\
\textcolor{black}{\textit{A1: What could happen to a paper if you leave it outside even if it does not move?\\
B1: If you leave it outside, it could get soggy and disintegrate in the rain.\\
A2: What if it's a sunny day? \\
B2: In that case, the sun could bleach the paper and make it brittle.}}

The Google evidence is:\textcolor{black}{\textit{ When paper is left outside, various environmental factors initiate complex chemical and physical transformations.Sunlight (UV) Exposure: Direct sunlight triggers photochemical deterioration, where ultraviolet (UV) radiation breaks chemical bonds in the paper's cellulose fibers. UV rays might break down color molecules (chromophores), leading to fading or bleaching. However, in most cases paper contains lignin, so it will yellow or turn brown due to oxidation.}} The fact-checker output is:\textcolor{black}{\textit{ \{'subclaims': [\{'text': 'Paper left outside can get soggy and disintegrate in the rain.', 'label': 'true', 'justification': 'The evidence states that exposure to rain leads to direct physical damage and excessive water fuels chemical reactions that weaken fiber bonds, leading to the paper becoming soggy or disintegrating.'\}, \{'text': 'Sunlight can bleach the paper and make it brittle.', 'label': 'true', 'justification': 'The evidence states that UV rays break down color molecules, leading to fading or bleaching, and the breakdown of molecular chains causes paper to lose flexibility and become brittle.'\}]\}}}

The fact-checker module ignored the part in the evidence mentioning that "However, in most cases paper contains lignin, so it will yellow or turn brown due to oxidation.", leading to labeling the second subclaim as true although it is false. This is a mis-interpretation/hallucination error (second error pattern) and it will cause the final verdict to be true although it should be false. This type of errors is not automatically detectable.

\subsection{How are Google AI mode generations validated}
\label{val_gg}
Google employs thousands of independent Search Quality Raters worldwide to provide a "human-in-the-loop" check: 
\begin{enumerate}
    \item E-E-A-T Framework: Raters evaluate content based on Experience, Expertise, Authoritativeness, and Trustworthiness. They manually flag content that feels "robotic," unoriginal, or factually incorrect.
    \item Side-by-Side Testing: Humans compare two different AI responses \footnote{\url{https://docs.cloud.Google.com/vertex-ai/generative-ai/docs/models/evaluation-overview}} to the same prompt and vote on which is more helpful and accurate to refine the underlying algorithms.
\end{enumerate}
In addition Google deploys several "live" systems work to maintain reliability:
\begin{itemize}
    \item Knowledge Graph Cross-Referencing: For factual queries, the system can cross-check claims against Google's structured database of known facts. \footnote{\url{https://www.linkedin.com/pulse/does-Googles-ai-lie-risks-ai-generated-search-results-ayub-ansary-q5r6c}}
    \item Source Attribution: To help you validate the information yourself, I provide citations and links to the primary sources I used to construct the answer.
    \item Feedback Loops: User signals—such as the "thumbs up/down" or "Report Incorrect Information" buttons—are used to identify and correct systemic errors over time. \footnote{\url{https://www.linkedin.com/pulse/does-Googles-ai-lie-risks-ai-generated-search-results-ayub-ansary-q5r6c}}
\end{itemize}
\label{Google_val}
\subsection{Experiments on DiaHalu}
\label{diaHalu_exp}

In this experiment, we extracted all factual and non-factual dialogues from the DiaHalu dataset and excluded other hallucination types, as they fall outside the scope of this study. In total, we obtained 245 instances, comprising 122 factual dialogues and 123 non-factual hallucinations. For a fair comparison, we adopted the prompt, settings and parameters from \citet{chen-etal-2024-diahalu} for hallucination detection and evaluated GPT-4 and GPT-5.2 on these 245 instances. We didn't perform any kind of data cleaning.

We report precision, recall, accuracy, as well as macro and weighted F1 scores. The results are presented in Table \ref{metric_CAAFC} for CAAFC, Table \ref{metric_GPT_52} for GPT-5.2, and Table \ref{metric_GPT_4} for GPT-4.

The results clearly show that, despite using the smaller quantized Gemma-27B model in CAAFC, the framework still outperforms substantially larger frontier models. This finding suggests that, for automated fact-checking (AFC) and non-factual hallucination detection, decomposing complex tasks into simpler sub-tasks can be more effective, allowing smaller LLMs to achieve superior performance.

We focused on examples where CAAFC correctly identifies the non-factual hallucination while GPT-4 and GPT-5.2 fails. We found out that subjective opinions projected as facts are sometimes misclassified as factual by GPT-5.2 and GPT-4. Here is the following non-factual conversation as an example:\textcolor{black}{\textit{
A1: Why would someone think about getting wet in the summer?\\
B1: Well, getting wet can actually keep you cool in the hot weather.\\ 
A2: But what about the inconvenience of being wet?\\
B2: True, but some people might enjoy the feeling of water on their skin. \\
A3: Is there any scientific reason behind this desire? \\
B3: Yes, the evaporation of water from your skin can help lower your body temperature. \\
A4: Are there any other benefits to getting wet in the summer?\\
B4: It can also be a fun way to engage in water activities and sports. \\
A5: What about the aesthetic appeal of water?\\
B5: Water can have a calming and soothing effect on the mind and body. \\
A6: Are there any cultural or literary references to getting wet in the summer?\\
B6: Yes, many poets and artists have romanticized the idea of getting wet in the summer rain. \\
A7: What about the nostalgia associated with water activities in the summer? \\
B7: Childhood memories of playing in the sprinklers or swimming at the beach can bring joy. \\
A8: How does the sensation of being wet enhance our sensory experience?\\
B8: The feeling of water on the skin can heighten our awareness of the environment.}}

The \textbf{dataset evidence }is: \textcolor{black}{\textit{no scientific basis for this view}}

\textbf{Google Evidence}:  \textcolor{black}{\textit{biophysics and psychology. Evaporative Cooling: This is a scientifically proven phenomenon. When water on your skin evaporates, it requires energy (heat) to change from liquid to gas. That energy is drawn directly from your body, effectively lowering your skin and core temperature. Calming Effects: The "Blue Mind" theory suggests that being near, in, or under water induces a meditative state. Research indicates this can lower heart rates, reduce cortisol (stress hormone) levels, and increase "feel-good" neurochemicals like dopamine and serotonin. Sensory Awareness: Water on the skin activates multisensory inputs—thermal (cold) and tactile (pressure/friction). While humans lack specific "wetness" receptors, our brains integrate these sensations to heighten our awareness of physical surroundings. 
The following from your conversation are subjective:
Enjoyment of the Sensation: The claim that "people might enjoy the feeling of water" is a subjective preference [B2]. While many do, others with sensory processing sensitivities (like some individuals with autism) may find the feeling of wet skin or wet clothes overwhelming or distressing.
Inconvenience of Being Wet: "Inconvenience" is an opinion [A2]. For some, the cooling benefit outweighs the mess; for others, the hassle of changing clothes makes it a net negative.
"Fun" of Water Activities: Labeling water sports as "fun" is an experiential claim [B4]. While they offer physical benefits, "fun" is an individual emotional response. }}

\textbf{Gemma3-27B output} is: \textcolor{black}{\textit{\{'subclaims': [\{'text': 'Getting wet can keep you cool in hot weather.', 'label': 'true', 'justification': 'The evidence states that evaporative cooling is a scientifically proven phenomenon where evaporating water draws heat from the body, lowering temperature.'\}, \{'text': 'people enjoy the feeling of water on their skin.', \textbf{'label': 'false'}, 'justification': 'The evidence states this claim is subjective and cannot be objectively proven.'\}, \{'text': 'Evaporation of water from your skin can help lower your body temperature.', 'label': 'true', 'justification': 'The evidence explicitly states that evaporation requires energy (heat) drawn from the body, lowering skin and core temperature.'\}, \{'text': 'Water activities and sports can be fun.', 'label': 'true', 'justification': "The evidence states that labeling water sports as 'fun' is an experiential claim and subjective."\}, \{'text': 'Water can have a calming and soothing effect on the mind and body.', 'label': 'true', 'justification': "The evidence mentions the 'Blue Mind' theory, which suggests being near water induces a meditative state, lowering heart rates and cortisol levels."\}, \{'text': 'Many poets and artists have romanticized the idea of getting wet in the summer rain.', 'label': 'true', 'justification': 'The evidence states that many authors, including Percy Shelley and Marguerite Duras, have used summer rain and water to symbolize various themes.'\}, \{'text': 'Childhood memories of playing in the sprinklers or swimming at the beach can bring joy.', 'label': 'true', 'justification': 'The evidence states that psychologists find that childhood memories of water are powerful emotional anchors that can trigger joy.'\}, \{'text': 'The feeling of water on the skin can heighten our awareness of the environment.', 'label': 'true', 'justification': 'The evidence states that water on the skin activates multisensory inputs, heightening our awareness of physical surroundings.'\}]\}}}
\textcolor{black}{\textit{
As we can see Gemma was able to spot the part that was projecting subjective opinions as facts. Another pattern seen is when the mis-information is presented in an indirect way in the conversation. For example: 
A1: I can't believe our wedding is just a month away! 
B1: I know, it feels like time flew by so quickly. Are you ready for the big day? 
A2: I think so, but there's still so much to do. Have you picked out your suit yet? 
B2: have you thought about our honeymoon destination? 
A3: I was thinking of a tropical beach paradise, what do you think?
B3: That sounds amazing, but I've always wanted to visit a European city as well. 
A4: How about we do both? Like Paris in France?
B4: Paris in France sounds incredibly romantic! It would be a perfect mix of a European city adventure and some relaxation on a tropical beach.}}

The conversation implies that Paris is a tropical beach paradise indirectly, so GPT-4 and GPT-5.2 didn't detect it. Here is the CAAFC using Gemma27B output:\textcolor{black}{\textit{
\{'subclaims': [\{'text': 'Paris is a tropical beach paradise.', 'label': 'false', 'justification': 'The evidence states Paris is not a tropical destination and does not have natural beaches.'\}, \{'text': 'Paris is a European city.', 'label': 'true', 'justification': 'The evidence confirms Paris is located in Europe.'\}, \{'text': 'Paris offers a mix of city adventure and relaxation on a tropical beach.', 'label': 'false', 'justification': 'The evidence states Paris cannot naturally offer a tropical beach experience.'\}]\}}}
The claim extractor-segmentor what makes CAAFC detects implicit misinformation and subjective opinions projected as facts.

\begin{table}
\centering
\resizebox{\columnwidth}{!}{%
\begin{tabular}{c|c|c|c|c|}
\cline{2-5}
 & \textbf{precision} & \textbf{recall} & \textbf{f1-score} & \textbf{support} \\ \hline
\multicolumn{1}{|c|}{\textbf{\begin{tabular}[c]{@{}c@{}}Class 0 \\ (factual)\end{tabular}}} & \textbf{0.877} & 0.467 & 0.610 & 122 \\ \hline
\multicolumn{1}{|c|}{\textbf{\begin{tabular}[c]{@{}c@{}}Class 1 \\ (hallucination)\end{tabular}}} & \textbf{0.639} & \textbf{0.935} & \textbf{0.759} & 123 \\ \hline
\multicolumn{1}{|c|}{\textbf{accuracy}} &  &  & \textbf{0.702} & 245 \\ \hline
\multicolumn{1}{|c|}{\textbf{macro avg}} & \textbf{0.758} & \textbf{0.701} & \textbf{0.684} & 245 \\ \hline
\multicolumn{1}{|c|}{\textbf{weighted avg}} & \textbf{0.757} & \textbf{0.702} & \textbf{0.685} & 245 \\ \hline
\end{tabular}%
}
\caption{Classification metrics report for CAAFC on the sampled DiaHalu dataset}
\label{metric_CAAFC}
\end{table}

\begin{table}
\centering
\resizebox{\columnwidth}{!}{%
\begin{tabular}{c|c|c|c|c|}
\cline{2-5}
 & \textbf{precision} & \textbf{recall} & \textbf{f1-score} & \textbf{support} \\ \hline
\multicolumn{1}{|c|}{\textbf{\begin{tabular}[c]{@{}c@{}}Class 0 \\ (factual)\end{tabular}}} & 0.853 & \textbf{0.475} & \textbf{0.611} & 122 \\ \hline
\multicolumn{1}{|c|}{\textbf{\begin{tabular}[c]{@{}c@{}}Class 1 \\ (hallucination)\end{tabular}}} & 0.638 & 0.919 & 0.753 & 123 \\ \hline
\multicolumn{1}{|c|}{\textbf{accuracy}} &  &  & 0.698 & 245 \\ \hline
\multicolumn{1}{|c|}{\textbf{macro avg}} & 0.746 & 0.697 & 0.682 & 245 \\ \hline
\multicolumn{1}{|c|}{\textbf{weighted avg}} & 0.745 & 0.698 & 0.682 & 245 \\ \hline
\end{tabular}%
}
\caption{Classification metrics report for GPT-5.2 the sampled DiaHalu dataset}
\label{metric_GPT_52}
\end{table}

\begin{table}
\centering
\resizebox{\columnwidth}{!}{%
\begin{tabular}{c|c|c|c|c|}
\cline{2-5}
 & \textbf{precision} & \textbf{recall} & \textbf{f1-score} & \textbf{support} \\ \hline
\multicolumn{1}{|c|}{\textbf{\begin{tabular}[c]{@{}c@{}}Class 0\\ (factual)\end{tabular}}} & 0.524 & 0.910 & 0.665 & 122 \\ \hline
\multicolumn{1}{|c|}{\textbf{\begin{tabular}[c]{@{}c@{}}Class 1\\ (hallucination)\end{tabular}}} & 0.667 & 0.179 & 0.282 & 123 \\ \hline
\multicolumn{1}{|c|}{\textbf{accuracy}} &  &  & 0.543 & 245 \\ \hline
\multicolumn{1}{|c|}{\textbf{macro avg}} & 0.595 & 0.544 & 0.473 & 245 \\ \hline
\multicolumn{1}{|c|}{\textbf{weighted avg}} & 0.595 & 0.543 & 0.473 & 245 \\ \hline
\end{tabular}%
}
\caption{Classification metrics report for GPT-4 the sampled DiaHalu dataset}
\label{metric_GPT_4}
\end{table}

\subsection{FACTors recent claims evaluation with LLAMA-CAAFC}
\label{af_ll}
The aim of this experiment is to evaluate the effect of using external evidence in mitigating potential LLM bias arising from internal knowledge or training data in AFC. We consider 196 claims from the FACTors dataset that were published on or after December 6, 2024—the release date of LLaMA 3.3–70B. We compare the accuracy and F1-score of LLaMA 3.3–70B on these 196 recent claims with its performance on older claims within FACTors. If similar performance is observed across both subsets, this would indicate that the model’s internal knowledge and potential training data bias do not significantly influence its predictions. Conversely, substantial performance differences would suggest an impact of internal knowledge on the model’s outputs. The performance metrics of LLAMA-CAAFC is shown in table \ref{after_llama}.

\begin{table}
\resizebox{\columnwidth}{!}{%
\begin{tabular}{c|c|c|c|c|}
\cline{2-5}
 & \textbf{precision} & \textbf{recall} & \textbf{f1-score} & \textbf{support} \\ \hline
\multicolumn{1}{|c|}{\textbf{\begin{tabular}[c]{@{}c@{}}class 1\\ (false)\end{tabular}}} & 0.988 & 0.918 & 0.952 & 182 \\ \hline
\multicolumn{1}{|c|}{\textbf{\begin{tabular}[c]{@{}c@{}}class 2\\ (true)\end{tabular}}} & 0.333 & 0.750 & 0.462 & 8 \\ \hline
\multicolumn{1}{|c|}{\textbf{\begin{tabular}[c]{@{}c@{}}class 3\\ (unverifiable)\end{tabular}}} & 0.556 & 0.833 & 0.667 & 6 \\ \hline
\multicolumn{1}{|c|}{\textbf{accuracy}} &  &  & 0.908 & 196 \\ \hline
\multicolumn{1}{|c|}{\textbf{macro avg}} & 0.626 & 0.834 & 0.693 & 196 \\ \hline
\multicolumn{1}{|c|}{\textbf{weighted avg}} & 0.948 & 0.908 & 0.923 & 196 \\ \hline
\end{tabular}%
}
\caption{Classification report for the claims that are published after the release date of LLAMA3.3-70B}
\label{after_llama}
\end{table}

When comparing the F1-score and accuracy of LLaMA 3.3–70B reported in Table \ref{framework_perf} and Table \ref{after_llama}, we observe that the results are nearly identical. This suggests that the model’s internal knowledge does not significantly influence its performance, as it primarily relies on the retrieved evidence as the reference for its veracity predictions. These finding are in alignment with \cite{rahman2026hallucination, russo2025face, yue2024retrieval}
\subsection{ Concern about the aggregation of data labels in fine-grained datasets}
\label{lab_agg}
Because CAAFC is evaluated on datasets that adopt heterogeneous labeling schemes, label aggregation is necessary to produce a unified and consistent output space across benchmarks. For example, labels such as misleading, partially true, and false are consolidated into a single false category. This harmonization is a methodological requirement rather than a design choice intended to favor CAAFC.

We acknowledge the concern that such aggregation might appear to create an unfair comparison with state-of-the-art (SOTA) baselines evaluated under the original fine-grained labels. However, the opposite scenario also occurs. In the case of CoverBench and the sampled DiaHalu—both contains only two labels (true or false)—CAAFC is capable of producing an additional unverifiable prediction. This introduces a structural disadvantage for CAAFC, as any unverifiable output is effectively penalized under the binary evaluation scheme. 

Despite this clear disadvantage, CAAFC still surpasses the SOTA baselines on CoverBench and the sampled DiaHalu (2 out of 4 datasets). This demonstrates that the observed performance gains cannot be attributed to label aggregation. 

Furthermore, we provide detailed justifications explaining why each claim is deemed true, false, or unverifiable, compensating for the lack of fine-grained classification. 
\subsection{ CAAFC hallucination and detection test on HaluEval}
\label{H_EV}

HaluEval \cite{li-etal-2023-halueval} is a knowledge-grounded dialogue hallucination benchmark with about 10,000 samples built by perturbing grounded responses from OpenDialKG and pairing them with gold/reference answers. The aim of this test is to confirm the capability of CAAFC to handle non-factual hallucination detection and correction.

The SOTA model of HaluEval dialogue dataset, ChatGPT , achieved an accuracy of 0.738  after incorporating relevant external knowledge. In comparison, CAAFC, utilizing the smaller Gemma 3-27B model, achieved an accuracy of 0.770.

There are 2196 instances that exhibit the first error pattern related to evidence mismatch due to an outdated evidence in the dataset or missing context, where all three LLMs identified 1998 of these responses as hallucinations, despite being labeled as factual in the dataset and 198 responses as factual despite being labeled hallucinations in the dataset. The rest of the 104 misclassifications are attributed to the second error pattern, which arises from the models mis-interpretations/hallucination. After data cleaning, the accuracy jumps to 0.986.

An example of the second error pattern where models interpretations are different is the following \textbf{conversation}:\textcolor{black}{\textit{
[Human]: Have you heard of a movie called Fury? Do you like it? [Assistant]: The movie was ok, the emotion is better.  Brad Pitt was pretty good.  Do you like movies about war that were converted from video games? [Human]: I like Brad Pitt. He starred in The Tree of Life Right? [Assistant]:"Yes he did.  Sean Penn and Fiona Shaw were also in it.  Do you like drama?"}}

\textbf{The Google evidence:}  Here is the information regarding the claims from your conversation:
\textcolor{black}{\textit{Fury (2014 Movie), Contrary to the suggestion of being a video game adaptation, Fury is a fictional narrative. Director David Ayer was inspired by the service of veterans in his family and books like Death Traps by Belton Y. Cooper.It stars Brad Pitt as Sgt. Don "Wardaddy" Collier, alongside Shia LaBeouf, Logan Lerman, Michael Peña, and Jon Bernthal. The assistant correctly identified Sean Penn (as Jack) and Fiona Shaw (as Grandmother) as cast members. The film also features Jessica Chastain.
Genre and Recognition: It is an epic experimental drama written and directed by Terrence Malick. It won the prestigious Palme d'Or at the Cannes Film Festival.}} 

Gemma Label is \textit{\textbf{"hallucination"}} as it assumed that the AI assistant implies that the movie was converted from a video game. Here is the output of Gemma: \textcolor{black}{\textit{\{'subclaims': [\{\textbf{'text': 'Fury was converted from a video game.', 'label': 'false', 'justification': 'The evidence states Fury is a fictional narrative and was not based on a video game, despite a cross-promotion with World of Tanks.}'\}, \{'text': 'Brad Pitt was in the movie Fury.', 'label': 'true', 'justification': "The evidence lists Brad Pitt as starring in Fury as Sgt. Don 'Wardaddy' Collier."\}, \{'text': 'Brad Pitt starred in The Tree of Life.', 'label': 'true', 'justification': "The evidence confirms Brad Pitt starred in The Tree of Life as Mr. O'Brien."\}, \{'text': 'Sean Penn was in The Tree of Life.', 'label': 'true', 'justification': 'The evidence confirms Sean Penn was in The Tree of Life as Jack.'\}, \{'text': 'Fiona Shaw was in The Tree of Life.', 'label': 'true', 'justification': 'The evidence confirms Fiona Shaw was in The Tree of Life as Grandmother.'\}]\}}}

However, GPT didn't judge the AI assistant based on any possible implications and it deemed the output as factual. Here is the GPT output:\textcolor{black}{\textit{\{'subclaims': [\{'text': 'Fury is a movie.', 'label': 'true', 'justification': 'The evidence provides information about Fury (2014 Movie), confirming it is a film.'\}, \{'text': 'Brad Pitt was in Fury.', 'label': 'true', 'justification': 'The evidence lists Brad Pitt as starring in Fury, playing Sgt. Don Wardaddy Collier.'\, \{'text': 'Brad Pitt starred in The Tree of Life.', 'label': 'true', 'justification': "The evidence states Brad Pitt did star in The Tree of Life as Mr. O'Brien."\}, \{'text': 'Sean Penn was in The Tree of Life.', 'label': 'true', 'justification': 'The evidence confirms Sean Penn appears in The Tree of Life, playing the character Jack.'\}, \{'text': 'Fiona Shaw was in The Tree of Life.', 'label': 'true', 'justification': 'The evidence lists Fiona Shaw as part of the cast of The Tree of Life, playing the Grandmother.'\}]\}}}

\subsection{ Human comparison between dataset evidence and Google evidence}
\label{hum_ann}
For evidence-quality evaluation, the framework uses three LLMs (Llama 3.3, Gemma 3-27B, GPT-OSS-120B) to decide when dataset labels/evidence are wrong and to prefer Google AI evidence. Then the same set of models is used as the fact-checker inside CAAFC, and performance is reported against the cleaned datasets that were curated using those LLMs. This may create a circular evaluation: LLMs are both judge and system under test. To mitigate this effect we perform an external human audit of the removed instances.

Three human annotators, MSc students familiar with NLP tasks, aged between 22 and 30 years, were asked to determine whether the Google AI–retrieved evidence is better than the original dataset evidence, following the criteria and guidelines shown in Figure \ref{inst_annot}. The annotators didn't know the sources of evidences or any related metadata. The annotators didn't receive any training. There are around 375 chronological mismatch instances in total across the datasets. The evaluation was conducted on 100 randomly selected examples exhibiting evidence–chronology mismatch patterns across the four combined datasets. The 100 examples include claims that were unanimously deemed by the 3 LLMs as true while the dataset label is false, and vice versa. One of these two evidences support the claim while the other refutes it. 

Annotators are instructed to assign the veracity label based solely on their selected preferred evidence. This requirement is intended to ensure careful attention and logical consistency throughout the annotation process. For example, if annotators select the first evidence as the preferred evidence because it supports the claim, yet assign a false veracity label, this indicates an internal inconsistency in their evaluation. Such a discrepancy suggests that the evidence assessment and the veracity judgment were not aligned, likely reflecting insufficient attention to the example.

In total, Annotator 1 exhibited 2 such inconsistencies, Annotator 2 exhibited 11, and Annotator 3 exhibited 7. Table \ref{hum_ann_corr} shows the Pearson correlation between the annotators. Since Annotator 2 exhibited the highest number of discrepancies, this annotator also shows the lowest correlation with the other two annotators.

\begin{figure}
\fcolorbox{blue}{gray!15}{%
        \begin{minipage}{\columnwidth}
\scriptsize

You are asked to perform two sequential tasks:
\begin{enumerate}
    \item \textbf{Evidence Comparison Task:} Compare two pieces of evidence (Evidence 0 and Evidence 1) related to the same claim and determine which one is better based on specific evaluation criteria mentioned later.
    \item \textbf{Claim Veracity Evaluation Task:} Using the preferred evidence from Task 1, determine the veracity of the claim.
\end{enumerate}
You must complete Task 1 first, then proceed to Task 2.\\\\
\textbf{Task 1 Evidence Comparison}\\
You will receive a claim, evidence 0, and evidence 1. Your goal is to determine which evidence is stronger and more reliable for fact-checking the claim\\
    \begin{itemize}
        \item Step 1: Evaluate Each Evidence Independently based on how recent and temporally appropriate the evidence is relative to the claim, how directly and sufficiently the evidence addresses the claim, and if it includes all the necessary details.
        \item Step 2: Choose your preferred evidence (evidence 0 or evidence 1)
    \end{itemize}
\textbf{Task 2 Claim Veracity Evaluation}\\
Use only the preferred evidence to determine the claim veracity (true, false):
\begin{itemize}
    \item True – The evidence clearly supports the claim.
    \item False – The evidence contradicts the claim.
\end{itemize}

 \end{minipage}%
}
    \caption{Instructions provided to the human annotators}
    \label{inst_annot}
\end{figure}

We also assess agreement among human annotators using inter-annotator agreement Krippendorff's Alpha for multi-rater reliability which reports 0.891. This high agreement is expected due to the following factors:
\begin{enumerate}
    \item Clear Annotation Guidelines – annotators have strict, well-defined rules. In addition, there is a direct channel (emails) for them to send any questions regarding the instructions.
    \item Objective or Easy-to-Classify Data – Tasks with minimal ambiguity (e.g., labeling with ’0, or ’ labels like "true and false") often lead to high agreement.
    \item Annotators with similar backgrounds tend to agree more than crowd-sourced annotators.
\end{enumerate}

A recruitment email was sent to postgraduate students and three annotators volunteered for this task. Given this, we can assert that their annotations were conducted solely based on their understanding of the provided instructions.

Next, we evaluate agreement between the majority vote of the human annotations and the unanimous labels produced by the LLMs. A majority vote across the three annotators is used to obtain a single human reference label, and agreement between this majority label and the LLM decision is measured using Cohen’s Kappa.

The Cohen’s Kappa for preferred evidence selection between the LLMs and the majority vote of the human annotators is 0.979, indicating significant agreement. For the veracity labels, Cohen’s Kappa is 0.946, which also reflects a very high level of agreement. The slight difference between these scores is mainly attributable to discrepancies and internal inconsistencies among the human annotators, particularly in cases where an annotator selected a piece of preferred evidence but assigned a contradictory veracity label.
\begin{table}
\resizebox{\columnwidth}{!}{%
\begin{tabular}{c|r|r|r|}
\cline{2-4}
\multicolumn{1}{l|}{} & \cellcolor[HTML]{FFFFFF}{\color[HTML]{1F1F1F} \textbf{ev\_ann1}} & \cellcolor[HTML]{FFFFFF}{\color[HTML]{1F1F1F} \textbf{ev\_ann2}} & \cellcolor[HTML]{FFFFFF}{\color[HTML]{1F1F1F} \textbf{ev\_ann3}} \\ \hline
\rowcolor[HTML]{F2F2F2} 
\multicolumn{1}{|c|}{\cellcolor[HTML]{F2F2F2}{\color[HTML]{1F1F1F} \textbf{ev\_ann1}}} & {\color[HTML]{1F1F1F} 1.000} & {\color[HTML]{1F1F1F} 0.898} & {\color[HTML]{1F1F1F} 0.921} \\ \hline
\rowcolor[HTML]{FFFFFF} 
\multicolumn{1}{|c|}{\cellcolor[HTML]{FFFFFF}{\color[HTML]{1F1F1F} \textbf{ev\_ann2}}} & {\color[HTML]{1F1F1F} 0.898} & {\color[HTML]{1F1F1F} 1.000} & {\color[HTML]{1F1F1F} 0.898} \\ \hline
\rowcolor[HTML]{F2F2F2} 
\multicolumn{1}{|c|}{\cellcolor[HTML]{F2F2F2}{\color[HTML]{1F1F1F} \textbf{ev\_ann3}}} & {\color[HTML]{1F1F1F} 0.921} & {\color[HTML]{1F1F1F} 0.898} & {\color[HTML]{1F1F1F} 1.000} \\ \hline
\end{tabular}%
}
\caption{Pearson correlation between the three evidence annotations of the three human annotators}
\label{hum_ann_corr}
\end{table}

\end{document}